\documentclass[11pt]{article}

\usepackage[utf8]{inputenc}
\usepackage[T1]{fontenc}
\usepackage{lmodern}
\usepackage[margin=1in]{geometry}
\usepackage{amsmath,amssymb}
\usepackage{graphicx}
\usepackage{booktabs}
\usepackage{xspace}
\usepackage{xcolor}
\usepackage{subcaption}
\usepackage[numbers,sort&compress]{natbib}
\PassOptionsToPackage{hyphens}{url}
\usepackage[colorlinks=true,allcolors=blue]{hyperref}
\usepackage{caption}
\usepackage{microtype} 

\graphicspath{{figures/}}

\IfFileExists{macros/results_macros.tex}{
\newcommand{\NminiPrimary}{300\xspace}
\newcommand{\NfullPrimary}{100\xspace}

\newcommand{\baseRejectRate}{50.0\xspace}
\newcommand{\aurocMini}{0.82\xspace}

\newcommand{\aurocMiniFull}{0.82 (95\% CI 0.78--0.87)\xspace}
\newcommand{\aurocFull}{0.87\xspace}

\newcommand{\aurocFullFull}{0.87 (95\% CI 0.79--0.93)\xspace}
\newcommand{\spearmanRatingMini}{0.52\xspace}

\newcommand{\spearmanRatingMiniFull}{0.52 (95\% CI 0.43--0.59)\xspace}

\newcommand{\spearmanRatingFullFull}{0.60 (95\% CI 0.45--0.71)\xspace}
\newcommand{\cohensDMini}{1.28\xspace}
\newcommand{\cliffsDeltaMini}{0.65\xspace}

\newcommand{\trendPrel}{p<0.0001\xspace}
\newcommand{\trendRho}{0.56\xspace}
\newcommand{\trendRhoCI}{[0.47, 0.63]\xspace}
\newcommand{\bottomRejectRate}{89.8\xspace}
\newcommand{\bottomRejectCI}{[79.5, 95.3]\xspace}
\newcommand{\bottomLift}{1.80\xspace}
\newcommand{\bottomOralRate}{3.4\xspace}

\newcommand{\bottomRejectRateFrontier}{100.0\xspace}
\newcommand{\bottomRejectCIFrontier}{[79.6, 100.0]\xspace}
\newcommand{\bottomLiftFrontier}{2.00\xspace}
\newcommand{\bottomOralRateFrontier}{0.0\xspace}
\newcommand{\bottomNFrontier}{15\xspace}
\newcommand{\bottomBandHiFrontier}{70\xspace}

\newcommand{\bridgeRhoFull}{0.81 (95\% CI 0.72--0.88)\xspace}
\newcommand{\bridgeN}{100\xspace}
\newcommand{\bottomLiftCI}{[1.63, 2.07]\xspace}

\newcommand{\bridgeLowRhoFull}{0.52 (95\% CI 0.06--0.80)\xspace}
\newcommand{\bridgeOverlap}{59.1\xspace}
\newcommand{\natAcceptRate}{27.4\xspace}
\newcommand{\natBaseReject}{72.6\xspace}
\newcommand{\natBottomPrecision}{95.7\xspace}

\newcommand{\aurocEqualWeightFull}{0.82 (95\% CI 0.77--0.87)\xspace}
\newcommand{\rhoEqualWeight}{0.92\xspace}
\newcommand{\looAurocMin}{0.79\xspace}
\newcommand{\looAurocMax}{0.83\xspace}

\newcommand{\aurocNoArxivPriorFull}{0.83 (95\% CI 0.77--0.87)\xspace}
\newcommand{\nArxivPrior}{70\xspace}
\newcommand{\naiveN}{100\xspace}

\newcommand{\aurocNaiveFull}{0.80 (95\% CI 0.70--0.88)\xspace}

\newcommand{\aurocDiffFullNaive}{0.07\xspace}
\newcommand{\aurocDiffCI}{[-0.01, 0.16]\xspace}

\newcommand{\aurocDiffPrel}{p=0.09\xspace}
\newcommand{\opSixtyFullAcc}{51.0\xspace}
\newcommand{\opSixtyNaiveAcc}{72.0\xspace}
\newcommand{\fullRunSD}{0.7\xspace}
\newcommand{\naiveRunSD}{2.8\xspace}
\newcommand{\relPairedP}{p=0.014\xspace}
\newcommand{\relPairedN}{10\xspace}
\newcommand{\lenRhoPage}{0.24\xspace}

\newcommand{\lenRhoWord}{0.20\xspace}

\newcommand{\lenRhoRefs}{-0.09\xspace}

\newcommand{\covAucMini}{0.87\xspace}
\newcommand{\scoreAucMini}{0.83\xspace}
\newcommand{\covAucFrontier}{0.83\xspace}
\newcommand{\scoreAucFrontier}{0.87\xspace}
\newcommand{\nCovMini}{299\xspace}
\newcommand{\nCovFrontier}{100\xspace}
\newcommand{\rhoWithinReject}{0.29\xspace}
\newcommand{\rhoWithinPoster}{0.03\xspace}
\newcommand{\rhoWithinOral}{0.10\xspace}
\newcommand{\rhoWithinAccepted}{0.11\xspace}
\newcommand{\bandSensMinReject}{87.2\xspace}
\newcommand{\bandSensMinLift}{1.74\xspace}
\newcommand{\bandSensNdefs}{5\xspace}
\newcommand{\disagreeRhoMini}{0.09\xspace}
\newcommand{\disagreeRhoFrontier}{0.11\xspace}
\newcommand{\disagreeAucLowMini}{0.86\xspace}
\newcommand{\disagreeAucHighMini}{0.79\xspace}
\newcommand{\disagreeAucLowFrontier}{0.91\xspace}
\newcommand{\disagreeAucHighFrontier}{0.84\xspace}
\newcommand{\nAreaRows}{14\xspace}
\newcommand{\areaAurocMin}{0.71\xspace}
\newcommand{\areaAurocMax}{1.00\xspace}
\newcommand{\costFullMiniTok}{223.8k\xspace}
\newcommand{\costFullTok}{287.9k\xspace}

\newcommand{\costFullVsMini}{1.3\xspace}
\newcommand{\runSDmedian}{0.7\xspace}
\newcommand{\runSDnPapers}{10\xspace}

\newcommand{\citationAUROC}{0.50\xspace}

\newcommand{\aurocDropCitationFull}{0.88\xspace}
\newcommand{\haloMeanR}{0.49\xspace}
\newcommand{\haloMinR}{0.30\xspace}
\newcommand{\haloMaxR}{0.73\xspace}
\newcommand{\haloN}{300\xspace}

}{}
\IfFileExists{macros/sim_macros.tex}{
\newcommand{\powerHoneN}{95\xspace}
\newcommand{\powerHtwoN}{100\xspace}

\newcommand{\mdeAUROC}{0.67\xspace}
\newcommand{\mdeAUROCdiff}{0.10\xspace}
\newcommand{\bootCoverage}{94\xspace}
\newcommand{\jtTypeOne}{2.7\xspace}
\newcommand{\jtUniformP}{0.26\xspace}
}{}
\IfFileExists{macros/secondary_macros.tex}{

}{}

\providecommand{\NminiPrimary}{\TBD}
\providecommand{\NfullPrimary}{\TBD}

\providecommand{\baseRejectRate}{\TBD}

\providecommand{\aurocMini}{\TBD}

\providecommand{\aurocMiniFull}{\TBD}
\providecommand{\aurocFull}{\TBD}

\providecommand{\aurocFullFull}{\TBD}

\providecommand{\spearmanRatingMini}{\TBD}

\providecommand{\spearmanRatingMiniFull}{\TBD}

\providecommand{\spearmanRatingFullFull}{\TBD}

\providecommand{\cohensDMini}{\TBD}
\providecommand{\cliffsDeltaMini}{\TBD}
\providecommand{\costFullVsMini}{\TBD}

\providecommand{\trendRho}{\TBD}
\providecommand{\trendRhoCI}{\TBD}

\providecommand{\bottomRejectRate}{\TBD}
\providecommand{\bottomRejectCI}{\TBD}
\providecommand{\bottomLift}{\TBD}
\providecommand{\bottomOralRate}{\TBD}

\providecommand{\bridgeN}{\TBD}
\providecommand{\runSDmedian}{\TBD}

\providecommand{\aurocNoArxivPriorFull}{\TBD}
\providecommand{\nArxivPrior}{\TBD}

\providecommand{\naiveN}{\TBD}

\providecommand{\aurocNaiveFull}{\TBD}

\providecommand{\opSixtyFullAcc}{\TBD}
\providecommand{\opSixtyNaiveAcc}{\TBD}
\providecommand{\fullRunSD}{\TBD}
\providecommand{\naiveRunSD}{\TBD}

\providecommand{\powerHoneN}{\TBD}
\providecommand{\powerHtwoN}{\TBD}

\providecommand{\mdeAUROC}{\TBD}
\providecommand{\bootCoverage}{\TBD}
\providecommand{\jtTypeOne}{\TBD}
\providecommand{\jtUniformP}{\TBD}

\newif\ifanonymous\anonymousfalse
\IfFileExists{ANONYMOUS.flag}{\anonymoustrue}{}

\title{\textbf{Intelligence Is Not the Bottleneck:\\
Validating an LLM First-Pass Manuscript Score Against Peer-Review Outcomes}}

\author{%
  \ifanonymous
  Anonymous (editor-blind build)\\
  \texttt{\small contact redacted}%
  \else
  Costa Georgantas\\
  \texttt{\small costa@aipr.pub}%
  \fi
}
\date{}

\begin{document}
\maketitle

\IfFileExists{macros/SYNTHETIC.flag}{%
  \begin{center}\fcolorbox{red}{red!8}{\parbox{0.9\linewidth}{\centering
  \textcolor{red}{\textbf{DRAFT ON SYNTHETIC DATA.} All numbers and figures below
  are generated from \texttt{synth.py} for formatting verification only and are
  not results. Re-run \texttt{run\_all.py} against the real export to populate.}}}\end{center}
}{}

\begin{abstract}
\noindent
Large language model (LLM) systems are increasingly proposed to assist peer review,
yet most evaluations judge the prose of machine-generated review \emph{text}, not the
validity of the numeric \emph{score} a system assigns. We validate \textsc{AIPR},
which reads a submitted manuscript and emits five 0--100 quality dimensions and a
weighted overall score, against the public decision outcomes of a major machine
learning venue. AIPR grades by prompting alone, with no fine-tuning on reviews or
decisions. Across \NminiPrimary{} ICLR submissions with public decision tiers and
reviewer ratings, graded under a frozen pipeline with hypotheses pre-registered before
any score met any outcome, the overall score separates rejected from accepted
submissions (AUROC~\aurocMiniFull), rises monotonically across tiers, and tracks the
mean reviewer rating. The signal is strongest where we claim it: the lowest-scoring
fifth is rejected far above the base rate, with oral papers absent. The validity comes mostly from the model: a one-paragraph prompt on
the same model discriminates almost as well as the full pipeline (the small gap favours
the pipeline but does not meet the pre-declared criterion, $p=0.09$). What the engineering adds is
\emph{reliability} and a grounded review: AIPR's score barely moves across repeated runs
(\fullRunSD{} vs.\ \naiveRunSD{} points within-paper SD) where the bare prompt swings,
and the same pass returns a rubric-structured, evidence-grounded review rather than a
bare number, with the human keeping the decision.
\end{abstract}

\section{Introduction}
\label{sec:intro}

Peer review is the gate through which scientific work passes, and it is under strain.
Submission volumes have grown faster than the reviewer pool, and the process is
demonstrably noisy: the NeurIPS consistency experiments found that a large fraction of
accept/reject decisions would flip under a re-run of the same process
\citep{cortes2021inconsistency,beygelzimer2023arbitrary}, with the asymmetry that
review is reliable at \emph{rejecting weak work} but poor at \emph{ranking strong
work}. Against this backdrop, LLM-based assistance is now being deployed at scale
\citep{thakkar2026feedback,biswas2026aaai}.

Most studies of automated review evaluate the \emph{text} a model produces: how well
its comments overlap a human reviewer's, how specific or factual they are
\citep{liang2024llm,darcy2024marg,yuan2022automate,du2024llmsassist}. For deployment
the more consequential question is whether the \emph{score} a system assigns is a valid
signal of the quality humans perceive. A score is what triages a queue or flags a
submission for a second look, so it must be validated as a measurement against an
external outcome.

\textsc{AIPR} is an interactive platform for reviewing manuscripts: it reads a submitted
PDF and returns a structured, evidence-grounded review, of which a weighted overall
quality score across five dimensions is one component (\S\ref{sec:methods}). The score is
an AI first-pass signal, not a final human-reviewed verdict; in deployment a reviewer
reads and acts on it interactively. AIPR grades by \emph{prompting alone}: it is not
fine-tuned on reviews, decisions, or ratings, so the score is a property of a prompted
frontier model, not of a classifier trained on outcomes. Here we isolate and benchmark
that first-pass score on first read, the hardest case for an automated signal and the one
a triage use relies on. Our ground truth is the public OpenReview record of a major
machine learning venue: each submission's decision tier (reject, poster, oral) and mean
reviewer rating; because the venue publishes \emph{rejected} submissions, the class our
central claim concerns is fully observable.

The paper turns on two pre-registered questions. \emph{Is the score valid?} That is, does
it agree with human outcomes well enough to flag weak submissions for a second look, the
low-end triage use we scope it to; we do not claim it predicts the acceptance decision,
ranks strong papers, or replaces a reviewer. \emph{What does the system add over the model
it runs on?} A frontier model is available to anyone, so we compare AIPR against a
one-paragraph prompt on the identical model and PDF, separating what the base model
contributes from what the engineered pipeline adds. The hypotheses and thresholds were
fixed before any score met any outcome (Appendix~\ref{app:prereg}); every number in the
paper is generated from the released data by one script, and the full study design is
diagrammed in Appendix~\ref{app:configs} (Fig.~\ref{fig:design}).

\section{Related Work}
\label{sec:related}

\paragraph{Automated review, and grading by prompting.}
Using LLMs to assist peer review has moved from speculation to deployment: a randomized
study at ICLR~2025 improved review informativeness with an LLM feedback agent
\citep{thakkar2026feedback}, and AAAI~2026 ran an AI-review pilot at scale
\citep{biswas2026aaai}. The anchor empirical study found GPT-4 review comments overlap
human comments about as much as two humans overlap \citep{liang2024llm}; multi-agent and
pipeline systems raise specificity \citep{darcy2024marg,lu2024aiscientist}; recent
benchmarks measure LLM reviewer ability and its divergence from humans
\citep{li2026llmreviewer,reviewertoo2025}; and critical work documents brittleness,
hidden-prompt attacks, and inflated scores
\citep{du2024llmsassist,zhou2024reliable,lin2025hidden,zhou2025positivereview,russolatona2024lottery,pangram2026iclr},
prompting funding agencies and journals to restrict undisclosed AI review
\citep{science2025funding}. Almost all of this evaluates generated review \emph{text}.
Separately, \emph{grading by prompting} is an established paradigm: strong LLMs score
open-ended text against rubrics with near-human agreement (LLM-as-a-judge;
\citealp{zheng2023judging,liu2023geval}), and prompt-only scoring has been validated
against human grades in automated essay scoring \citep{mizumoto2023exploring}. We build
on that paradigm rather than claim it.

\paragraph{Noisy ground truth, and how to validate against it.}
Any claim that a signal tracks ``quality'' must be calibrated against the unreliability
of the human ground truth. The NeurIPS consistency experiments are canonical: roughly
half of accepted papers would change under a re-run, and review succeeds at rejecting
weak papers but struggles to rank strong ones
\citep{cortes2021inconsistency,beygelzimer2023arbitrary}; bias and its measurement are
well documented \citep{tomkins2017reviewerbias,stelmakh2019biases,shah2022challenges}.
For validating a score against noisy, ordinal outcomes the field uses rank correlation,
threshold-free AUROC, and calibration assessment \citep{spearman1904,guo2017calibration},
on corpora such as PeerRead and NLPeer \citep{kang2018peerread,dycke2023nlpeer};
\citet{checco2021aiassisted} validated a learned score against decisions and ratings,
though with a trained classifier rather than an LLM.

\paragraph{Positioning.}
The older paradigm for scholarly assessment \emph{trains} on reviews and decisions
(acceptance prediction; \citealp{kang2018peerread}) or \emph{fine-tunes} review
generators \citep{idahl2025openreviewer}. Prompt-only treatment of manuscripts has so
far targeted feedback \emph{text} \citep{liang2024llm} or has been validated only against
weak proxies, such as an author's own ratings of his own papers
\citep{thelwall2024chatgpt}. To our knowledge this is the first study to validate a
training-free, prompt-only manuscript quality \emph{score} against the public \emph{reject}
class and reviewer ratings, and to decompose how much of that validity comes from the
underlying model versus the engineered pipeline. The individual ingredients each have
precedent: prompting, pre-registration, and outcome validation are all established. What
is new is their combination on the observable reject class, together with the
model-versus-pipeline reliability finding. The novelty is the validity result, not the
prompting.

\section{Data}
\label{sec:data}

\paragraph{Venue and ground truth.}
We draw submissions from a recent edition of a major machine learning conference on
OpenReview (primary cohort: ICLR~2026), chosen for two properties. It operates open
review, so \emph{rejected} submissions, their reviews, and decisions are all public;
venues that publish only accepted papers cannot support a claim about the reject class.
And its decisions and reviews were released in January~2026, after the grading model's
August~2025 knowledge cutoff \citep{openai2026gpt54}, so the outcome could not have been
memorized (\S\ref{sec:discussion}). For each submission we record two ground-truth signals: the
\textbf{decision tier} on the ordered ladder reject $<$ poster $<$ oral (and its binary
collapse to accept/reject), and the \textbf{mean reviewer rating}, a continuous signal
that partially averages out individual reviewer noise. We grade the \textbf{submitted}
PDF a reviewer actually saw, never the camera-ready, so the score is computed on the
same artifact the human signal responded to, and we record its SHA-256 for provenance.

\paragraph{Sampling, exclusions, and nested cohorts.}
The natural acceptance rate would starve the accepted tiers, so we sample
\emph{balanced} across decision tiers and report metrics both as sampled and re-weighted
to the venue's natural prevalence (\S\ref{sec:results}, Appendix~\ref{app:robust});
Table~\ref{tab:sample} gives the realized counts. Pre-registered exclusions
(desk-rejected or withdrawn submissions with no reviewer signal, PDFs that failed to
parse, and arXiv-twin leakage risks flagged by the pipeline's self-identity step) never
enter a primary metric. To establish the relationship at scale under a fixed budget,
gradings are organized into two strictly nested cohorts: every sampled submission is
graded by AIPR on the cheaper GPT-5.4-mini (cohort $M$, $n=\NminiPrimary$), and a
stratified subset is additionally graded by AIPR on the frontier GPT-5.4 (cohort
$H \subseteq M$). The two differ only in the model, so each frontier grading has a
paired cheap-model score (the bridge, H5).

\begin{table}[t]
  \centering
  \caption{Realized sample. Cohort $M$ (GPT-5.4-mini) carries the statistical power; the
  nested cohort $H$ (GPT-5.4) carries the production model. Both run the full two-pass
  pipeline. Counts exclude pre-registered exclusions.}
  \label{tab:sample}
  \begin{tabular}{lr}
\toprule
Decision tier & Submissions (cohort M) \\
\midrule
Reject & 150 \\
Poster & 100 \\
Oral & 50 \\
\midrule
\textbf{Total (cohort M, GPT-5.4-mini)} & \textbf{300} \\
AIPR (GPT-5.4), cohort H ($\subseteq$ M) & 100 \\
\bottomrule
\end{tabular}

\end{table}

\paragraph{Data contract and release.}
All analysis reads a two-table contract (one row per submission for labels, one per
grading run for scores; schema in Appendix~\ref{app:schema}), released with the analysis
code so every figure and number reproduces with one command. The released tables,
anchored to the public pre-registration tag, \emph{are} the cohort manifest a reader
audits against. We do not release the full AIPR review text for every identifiable
submission by default; that corpus is a controlled audit artifact available to editors
or reviewers on request.

\paragraph{Data licensing.}
All decision outcomes, reviewer ratings, and review excerpts are drawn from the public
OpenReview record of ICLR~2026 (\texttt{ICLR.cc/2026/Conference}). Per the OpenReview
Terms of Use, reviews, meta-reviews, and decisions are released under the Creative
Commons Attribution 4.0 International license (CC~BY~4.0), and each ICLR~2026 note
carries this license in its public metadata; review excerpts are reproduced under that
license with attribution to their anonymous reviewers via the corresponding OpenReview
forums. We redistribute only derived metadata (forum identifiers, decision tiers, mean
ratings) and our own scores; we do not redistribute manuscript PDFs or full review
texts.

\section{Methods}
\label{sec:methods}

\subsection{The scoring system}
\textsc{AIPR} grades a manuscript in a single frozen pipeline version (v6), applying the
now-standard practice of scoring by prompting an LLM against a rubric
\citep{liu2023geval,zheng2023judging} to a new object: the manuscript. The pipeline
uploads the submitted PDF and runs two passes. A \emph{reviewer} pass reads the paper
and emits the five dimension scores with justifications and verbatim-anchored
highlights. An \emph{audit} pass re-reads the paper with a live literature-search tool,
checks each highlight against the manuscript, and proposes missing citations against a
real bibliographic index, so a recommended citation is always a record the tool
returned, not a generated string. The overall score is a fixed weighted mean of the five
dimensions,
\begin{equation}
  \text{overall} = \frac{4\,s_{\text{nov}} + 2\,s_{\text{rig}}
  + 4\,s_{\text{app}} + 1\,s_{\text{cla}} + 0.5\,s_{\text{cit}}}{11.5},
  \label{eq:overall}
\end{equation}
with novelty and applicability weighted most and citation least; the model does not emit
the overall directly. The pipeline's product is this structured, evidence-grounded
review; the overall score is a by-product of producing it, not the system's purpose, and
the baseline of \S\ref{sec:res-value} returns only the number. The system is prompt-only:
it is not fine-tuned on reviews, decisions, or ratings, and no model here is trained on
the evaluation data.

The contrast with the baseline is concrete. Where the baseline prompt
(\S\ref{sec:res-value}) is one paragraph of about fifty words, the reviewer pass is
driven by a ${\approx}2{,}200$-word rubric instruction paired with a $1{,}200$-word
output schema, and the audit pass adds ${\approx}1{,}200$ words plus a tool loop of up to
eight OpenAlex search rounds. The two passes are role-separated, opening respectively
``You are an expert peer reviewer\,\ldots\ Produce a structured review'' and ``You are a
senior scientific editor\,\ldots\ audit the reviewer's highlights against the paper, and
find real missing citations.'' The full rubric, anchors, and schema are proprietary
(\S\ref{sec:discussion}); the point of \S\ref{sec:res-value} is that this instruction
footprint, almost two orders of magnitude larger than the baseline, buys reliability and
a grounded review rather than a higher discrimination score.

\subsection{Configurations and the baseline}
Two configurations run the identical two-pass pipeline and differ only in the model:
\textbf{AIPR (GPT-5.4-mini)} on the cheaper model and \textbf{AIPR (GPT-5.4)} on the
frontier model, the production configuration.\footnote{Exported as \texttt{full\_mini}
and \texttt{full\_full} in the released \texttt{gradings.csv}; we use the display names
throughout. The baseline \textbf{Direct (GPT-5.4)} is exported as \texttt{naive}.} The
cheap configuration is effectively unconstrained in volume; the frontier configuration is
budgeted. We therefore establish the score--outcome relationship on the large cohort $M$
and confirm it on cohort $H$, after verifying the two scores agree well enough for the
former to proxy the latter (H5). The pre-registered baseline, \textbf{Direct (GPT-5.4)},
runs the same frontier model on the same PDF with a single one-paragraph prompt and no
rubric or audit. We treat the system as a fixed black box: the dimension weights were set
by empirical internal validation on a separate, held-out manuscript set during product
development (no ICLR submissions from any year, none of the papers, decisions, or ratings
used here) and frozen before this study, not fit to any outcome in this cohort. The
lightly weighted citation dimension was uninformative in this run for a technical reason
in its audit, so we exclude it from interpretation; the headline is unchanged when it is
dropped, and a grounded, search-backed citation signal is an obvious next step
(\S\ref{sec:discussion}). We evaluate the rule \emph{as deployed} and report weighting
sensitivity (equal-weight, leave-one-out) in Appendix~\ref{app:robust}.

\subsection{Hypotheses, metrics, and pre-registration}
All hypotheses, the primary metric, and the low-score threshold were fixed before any
score was joined to any outcome (Appendix~\ref{app:prereg}). We report effect sizes with
bootstrap confidence intervals throughout, because ``strong signal'' is an effect-size
claim rather than a $p$-value claim. H1--H4 assemble criterion-related validity evidence
for a specific interpretation (weak relative to the venue bar) under a specific use
(triage), following the \emph{Standards for Educational and Psychological Testing}
\citep{aera2014standards}; a separate comparison (V1) asks whether that validity is the
pipeline's or the model's.

\begin{description}\itemsep2pt
  \item[H1 (primary): low-end flagging.] Submissions in the lowest score quintile are
  rejected well above the base rate, with oral papers essentially absent. Reported as
  reject rate and \emph{lift} over the base rate, with Wilson intervals.
  \item[H2: separation.] The overall score discriminates reject from accept (AUROC,
  stratified bootstrap CI).
  \item[H3: monotone gradient.] Mean overall score increases across the three ordered
  tiers (Jonckheere--Terpstra trend test, Monte Carlo permutation $p$; Spearman $\rho$).
  \item[H4: continuous agreement.] The overall score correlates with the mean reviewer
  rating (Spearman $\rho$, bootstrap CI).
  \item[H5 (gate): mini-to-frontier bridge.] On cohort $H$, AIPR (GPT-5.4-mini) and
  AIPR (GPT-5.4) overall scores correlate at $\rho \ge 0.8$, licensing the large-$N$
  cohort-$M$ results as a proxy for the production score.
  \item[V1 (primary value): pipeline vs.\ baseline.] On cohort $H$, AIPR (GPT-5.4)
  out-discriminates Direct (GPT-5.4) on the same model and PDF. Pre-declared success:
  AUROC difference $>0$ with the CI excluding zero; reported alongside accept/reject
  agreement and run-to-run reliability. A null is reported plainly.
\end{description}
Confidence intervals are 4{,}000-resample BCa bootstraps (stratified for class-conditional
statistics such as AUROC; Appendix~\ref{app:power}); the trend test uses 10{,}000 Monte
Carlo permutations; multiple comparisons across the five dimensions use Benjamini--Hochberg;
and the analysis is deterministic under a fixed seed. Secondary analyses (per-dimension
discrimination, audit-pass contribution, run-to-run variance, prevalence re-weighting,
weighting robustness, leakage controls) are reported in \S\ref{sec:results} and
Appendix~\ref{app:robust}.

\paragraph{Deviations from pre-registration.}
Two estimator choices differ from the registered analysis plan. Both are
analysis-stage substitutions of a better-behaved interval for the registered
one, and neither alters a pre-declared success threshold. First, confidence
intervals are BCa bootstraps rather than the registered percentile bootstrap:
the estimator-validation simulation (Appendix~\ref{app:power}) shows the
percentile interval under-covers at this sample size while BCa attains nominal
coverage. The V1 paired AUROC-difference CI is the exception and remains a
percentile interval on the paired bootstrap distribution. Second, the H1
success rule (lift CI lower bound above 1) reads a whole-cohort bootstrap CI
for the bottom-band lift rather than the registered Wilson-based interval:
resampling the whole cohort propagates uncertainty in the base rate and in the
sample-relative band edges, which a Wilson interval on the band's reject rate
alone cannot capture. Wilson intervals are still reported descriptively
alongside the bands (Table~\ref{tab:bands}).

\section{Results}
\label{sec:results}

The two pre-registered questions are not on equal footing. The first is the headline:
does the AIPR score agree with the human outcome (\S\ref{sec:res-validity}, H1--H5)?
Yes, and that agreement licenses the triage claim regardless of the second question. The
second asks what the engineered pipeline adds over the model alone
(\S\ref{sec:res-value}, V1). We lead on the large cohort $M$ ($n=\NminiPrimary$) and
confirm on cohort $H$ ($n=\NfullPrimary$); Table~\ref{tab:headline} gives both.

\subsection{The score agrees with the human outcome}
\label{sec:res-validity}

\begin{table}[t]
  \centering
  \caption{Headline metrics on the primary venue, for AIPR (GPT-5.4-mini) (statistical
  power) and AIPR (GPT-5.4) (production). Brackets are 95\% bootstrap confidence
  intervals.}
  \label{tab:headline}
  \begin{tabular}{lcc}
\toprule
Metric & AIPR (GPT-5.4-mini) ($n=300$) & AIPR (GPT-5.4) ($n=100$) \\
\midrule
AUROC (reject vs.\ accept) & 0.82 [0.78, 0.87] & 0.87 [0.79, 0.93] \\
Spearman $\rho$ (reviewer rating) & 0.52 [0.43, 0.59] & 0.60 [0.45, 0.71] \\
Cohen's $d$ (accept$-$reject) & 1.28 & 1.46 \\
Cliff's $\delta$ & 0.65 & 0.74 \\
Trend $\rho$ (across tiers) & 0.56 & 0.69 \\
Trend test $p$ & <0.0001 & --- \\
\bottomrule
\end{tabular}

\end{table}

\paragraph{Weak submissions are flagged (H1).}
We report the deployable flag on the production model, AIPR (GPT-5.4), directly. Its
lowest score quintile (the bottom fifth of the \NfullPrimary{}-paper cohort $H$;
integer-score ties shrink the strict quantile band to $n=\bottomNFrontier$) is rejected
at \bottomRejectRateFrontier\% (95\% CI \bottomRejectCIFrontier),
a \bottomLiftFrontier$\times$ lift over the \baseRejectRate\% base rate, with no oral
papers in that band (\bottomOralRateFrontier\%); the small band makes cohort $M$ the
anchor for scale. We use the
production score here because the mini-to-frontier bridge is weakest at the low end where
the flag fires (H5), so computing the flag on AIPR (GPT-5.4) sidesteps the proxy. The
pattern holds at scale on cohort $M$ (Table~\ref{tab:bands}): the lowest band is rejected
at \bottomRejectRate\% (95\% CI \bottomRejectCI), a \bottomLift$\times$ lift whose CI
\bottomLiftCI{} clears the pre-registered rule, with oral papers \bottomOralRate\% of the
band. The reliability curve (Appendix Fig.~\ref{fig:validation}b) falls monotonically as
the score rises. At the venue's natural prevalence (\natAcceptRate\% accept) the
bottom-quintile reject precision is \natBottomPrecision\%.

\begin{table}[t]
  \centering
  \caption{Reject rate, lift over base rate, and oral-paper share by AIPR-score quintile
  (cohort $M$). Band Q1 is the lowest-scoring fifth.}
  \label{tab:bands}
  \begin{tabular}{lccccc}
\toprule
Band & $n$ & Score range & Reject rate [95\% CI] & Lift [95\% CI] & Oral rate \\
\midrule
Q1 & 59 & 44--63 & 90\% [80, 95] & 1.80 [1.63, 2.07] & 3\% \\
Q2 & 52 & 64--66 & 77\% [64, 86] & 1.54 [1.29, 1.74] & 4\% \\
Q3 & 56 & 67--69 & 45\% [32, 58] & 0.89 [0.65, 1.16] & 16\% \\
Q4 & 49 & 70--71 & 35\% [23, 49] & 0.69 [0.41, 0.93] & 18\% \\
Q5 & 84 & 72--81 & 18\% [11, 27] & 0.36 [0.18, 0.51] & 33\% \\
\bottomrule
\end{tabular}

\end{table}

\paragraph{Separation, gradient, and rating agreement (H2--H4).}
The overall score separates reject from accept at AUROC \aurocMiniFull on cohort $M$,
confirmed at \aurocFullFull on cohort $H$ (ROC in Fig.~\ref{fig:naive}a); the
accepted--rejected gap is large (Cohen's $d=\cohensDMini$, Cliff's
$\delta=\cliffsDeltaMini$). Mean score rises monotonically across the three tiers
(Jonckheere--Terpstra $\trendPrel$; Spearman $\rho=\trendRho$, 95\% CI \trendRhoCI). And
it tracks the continuous mean reviewer rating at $\rho =$~\spearmanRatingMiniFull{} (cohort
$M$) and \spearmanRatingFullFull{} (cohort $H$); because the rating averages over reviewer
noise, this is the cleaner test, and the effect is strong rather than merely significant.

\ifanonymous
The agreement shows plainly at the score range's ends: AIPR's top-scoring submission was
an accepted oral (overall~83), its lowest a reject (overall~58, rating~3.5).
\else
The agreement shows plainly at the score range's ends: AIPR's top-scoring oral was
\citet{astabench2026} (overall~83), a public ICLR~2026 submission named for
concreteness; its lowest-scoring submission was a reject (overall~58, rating~3.5),
left unnamed as a courtesy to its authors.
\fi

\paragraph{The cheap-model proxy is valid (H5).}
On cohort $H$, AIPR (GPT-5.4-mini) and AIPR (GPT-5.4) correlate at $\rho =$~\bridgeRhoFull{}
($n=\bridgeN$; Fig.~\ref{fig:bridge}), clearing the pre-registered $\rho\ge0.8$ gate.
Because a high global correlation can hide low-end disagreement, we check directly:
within the cohort-$M$ bottom quintile the two still correlate at $\rho =$~\bridgeLowRhoFull{},
and \bridgeOverlap\% of that band is also in the frontier bottom quintile. Cohort $M$
therefore stands in for the production score, with cohort $H$ as confirmation.

\subsection{What the engineering adds: a strong model, made reliable}
\label{sec:res-value}
A frontier model is available to anyone, so what does the pipeline add over asking it
directly? The pre-registered value comparison (V1) grades cohort $H$ a second way,
Direct (GPT-5.4): the same model and PDF with one paragraph,
\begin{quote}\itshape
I'm reviewing this paper for ICLR, a top machine learning conference. Read it and tell me
how good it is: give a single overall quality score from 0 to 100 (where higher means
more likely to deserve acceptance) and a one or two sentence reason for the score.
\end{quote}
\noindent no rubric, audit, or examples, so the baseline is realistic rather than a
strawman.

\begin{figure}[t]
  \centering
  \includegraphics[width=\textwidth]{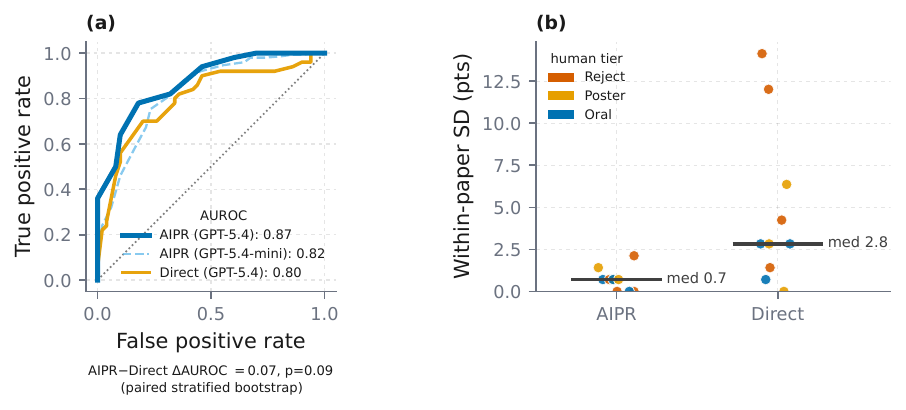}
  \caption{\textbf{What the engineering adds: a strong model made reliable} (cohort $H$;
  pre-registered comparison V1, same model and submitted PDF throughout). \textbf{(a)}
  Reject-vs-accept ROC for AIPR (GPT-5.4) (lead), AIPR (GPT-5.4-mini), and the Direct
  (GPT-5.4) baseline; all three clear the diagonal. The legend gives each grader's AUROC
  (95\% CIs in Table~\ref{tab:headline}); the AIPR$-$Direct paired difference (stratified
  bootstrap, $n=\naiveN$) is not statistically resolved (\aurocDiffPrel), so the
  engineering is not buying raw discrimination. \textbf{(b)} Where it pays off: within-paper score SD across repeated
  runs, one dot per variance-study paper coloured by human decision tier, with a median
  bar. AIPR grades far more consistently (median \fullRunSD{} vs.\ \naiveRunSD{} points),
  a gap that is large where the discrimination gap is not. Per-tier distributions:
  Appendix~\ref{app:robust}, Fig.~\ref{fig:score-dist}. Interactive counterpart:
  \url{https://aipr.pub/publications\#fig-naive}.}
  \label{fig:naive}
\end{figure}

\paragraph{The model already carries the signal.}
Direct (GPT-5.4) separates reject from accept on its own at AUROC \aurocNaiveFull{}. AIPR
(GPT-5.4) reaches \aurocFullFull{} on the same papers; the paired difference is small and
favours the pipeline but does not meet the pre-declared superiority criterion
(\aurocDiffFullNaive, 95\% CI \aurocDiffCI, \aurocDiffPrel; $n=\naiveN$). The direction is
consistent with a pipeline advantage and the result is suggestive rather than null. A
post-hoc power analysis (Appendix~\ref{app:power}) puts the smallest paired AUROC gap
detectable with 80\% power at this $n$ near \mdeAUROCdiff, above the observed
\aurocDiffFullNaive, so we claim neither superiority nor equivalence (ROC,
Fig.~\ref{fig:naive}a). The validated agreement of \S\ref{sec:res-validity} is therefore
not manufactured by the engineering; with or without the rubric, a frontier model produces
a first-pass score that tracks the human outcome, and raw intelligence is not the
bottleneck. The open question is \emph{where} the signal comes from, not \emph{whether} it
is there.

\paragraph{What the pipeline adds is reliability.}
The engineering buys not a higher score but a stable one. Across repeated gradings of the
same paper, AIPR's overall barely moves: median within-paper SD \fullRunSD{} points,
against \naiveRunSD{} for Direct on the same model and papers (\runSDnPapers{} papers,
each SD estimated from two consistent-configuration re-runs; Appendix~\ref{app:robust};
Fig.~\ref{fig:naive}b). The gap is systematic across papers, not only a median
difference: pairing the within-paper SDs of the two graders gives an exact Wilcoxon
signed-rank \relPairedP{} ($n=\relPairedN$ pairs). Where the discrimination gap is unresolved, this reliability gap
is large and unambiguous: the same submission earns nearly the same verdict every time,
and that run-to-run noise is small against the between-tier separation the score must
express. A reproducible score is worth more to a reviewer than one marginally sharper on
average but swinging several points between identical runs. And the same pass returns the
structured review itself (rubric-anchored dimensions, a grounded literature audit,
verbatim-anchored comments), which is the system's actual output; the score is a
by-product, and Direct returns only the number. That review, not the scalar, is what a
reviewer acts on. The pre-registered AIPR@60 cutoff is a diagnostic, not the deployment claim: the
production score is compressed high near the venue bar (\S\ref{sec:discussion}), so a
fixed 60 threshold predicts almost every paper accept (\opSixtyFullAcc\% vs.\
\opSixtyNaiveAcc\% balanced accuracy for AIPR and Direct) and we do not read it as a
classifier.

\subsection{Secondary analyses and failure modes}
\label{sec:res-secondary}
Per-dimension discrimination across the four informative dimensions
(Appendix~\ref{app:robust}, Fig.~\ref{fig:subscore-auroc}) is comparable across all
four, with rigor and applicability highest and novelty lowest. The
lightly weighted citation dimension is excluded from interpretation (uninformative this
run; \S\ref{sec:discussion}), and the headline is unchanged when it is dropped (AUROC
\aurocDropCitationFull{} vs.\ \aurocFull{}). Audit-pass contribution, prevalence
re-weighting, and weighting robustness are in Appendix~\ref{app:robust}; a second-venue
replication is future work.

\paragraph{Where the score errs.}
A triage signal must be characterized where it fails. Reading the public reviews of
cohort $H$ (full case studies, Appendix~\ref{sec:failure}), the dangerous error for
triage is a weak paper scored high: these concentrate where rejection rests on a
contribution-level judgment a competent manuscript surface does not betray, or on a
defect the system cannot see (it cannot run an experiment or re-derive a proof). The
mirror error, strong work scored low, reflects a genuine concern a reviewer also raised
but the community forgave, and costs only a second human look. The asymmetry is what
makes the narrow ``flag weak work'' claim defensible: false negatives at the top of the
range are quantified directly in the reliability curve, and false positives cost the
second look that is the intended action anyway.

\section{Discussion}
\label{sec:discussion}

\paragraph{Intelligence is not the bottleneck; reliability is.}
The load-bearing result is that Direct (GPT-5.4), a one-paragraph prompt, already
separates rejected from accepted submissions on its own (AUROC \aurocNaiveFull),
indistinguishably from AIPR ($\Delta$AUROC \aurocDiffFullNaive, \aurocDiffPrel). Raw
judgment is not what this deployment is shortest of; reliability is, the bare prompt's
score swinging \naiveRunSD{} points between identical runs against AIPR's \fullRunSD{}.
The open problem is not building a model that \emph{can} judge but turning that
capability into an instrument a reviewer can rely on: one that returns the same verdict
across runs, grounds its claims in evidence, and surfaces an anchored review instead of a
number. That is an engineering problem, and the one AIPR addresses. The field's effort is
better spent on the processing layer (reliability, grounding, interaction) than on a more
intelligent base model the task does not appear to need.

\paragraph{Compressed scores are a feature, not a defect.}
AIPR's production scores are compressed: they cluster high, near the venue bar, and move
little between runs (within-paper SD \fullRunSD{} vs.\ \naiveRunSD{} for the bare prompt).
Such clustering is a known behaviour of scalar LLM judges, which compress toward the top
of the scale and are unstable under resampling \citep{girrbach2025referencefree}, and for
a deployed grader reliability rather than spread is the property that governs whether the
score can be trusted \citep{gu2024surveyllmjudge,schroeder2024trustllm}. We trade spread
for stability deliberately: a stable estimate near a reference value is easier to act on
than a wider-ranging but noisier one. The reference is concrete in this cohort: the
production score's bottom quintile tops out at \bottomBandHiFrontier{} points and is
rejected at \bottomRejectRateFrontier\%, so scores clustering just above
\bottomBandHiFrontier{} sit near the venue's effective accept bar. The alternative, a
low point estimate carried with a wide interval, fails as a deployment object because a
reader anchors on the point estimate and under-weights the uncertainty around it
\citep{hofman2020inferential}; a stable scalar is the safer thing to hand a reviewer.
Because a quality score is subjective and reference-relative, a compressed, anchored value
also communicates where a submission sits against the venue bar without an explicit
pairwise comparison. We do not claim compression improves discrimination (AIPR's AUROC is
not higher than the bare prompt's), nor that it improves reviewers' actual decisions,
which we have not measured and which would need a calibration study or a user trial; the
claim is about the estimator and how it is read, not measured downstream behaviour. On
that bounded basis, a stable compressed scalar is a better-behaved object to act on.

\paragraph{An obvious next step: grounded citation checking.}
The deployed score includes a lightly weighted citation dimension that was uninformative
in this run for a technical reason in its audit and is excluded from the analysis here
(the headline is unchanged without it). Adding a grounded, search-backed citation signal,
which looks references up against real records rather than generating them, is the natural
next addition and the same reliability lesson as our headline; it matters given fabricated
references in accepted papers at NeurIPS~2025 \citep{neurips2025citations}. We omit it here
rather than report a known-broken dimension.

\paragraph{Do the subscores reflect independent assessment?}
A multi-dimensional score invites the halo question: does the model judge each dimension
on its merits or fit the parts to a global verdict? The four informative dimensions are
moderately correlated (mean $r=\haloMeanR$; Table~\ref{tab:subscore-corr}), sharing a
general-quality component but keeping distinct variance, so the breakdown is not one
number wearing four hats, though the correlation is consistent with partial anchoring.
Eliciting each dimension independently is the natural way to decorrelate the verdict from
the parts; we leave that comparison, with an order-perturbation probe, to future work.

\paragraph{Relation to work on review text and decision prediction.}
ReviewBench \citep{reviewbench2026} scores review \emph{comments} by structural
completeness and, like us, frames AI as complementary, but measures intrinsic text
properties, ``not their correctness,'' where we measure the extrinsic validity of a
\emph{score} against an outcome. A separate line predicts the accept/reject
\emph{decision}: from review text and sentiment
\citep{wang2018sentiment,ghosal2019deepsentipeer,kang2018peerread,jung2025whatdrives},
and recently from the manuscript itself with prompted LLM reviewers at near-human accuracy
\citep{reviewertoo2025}. That last result corroborates our headline, that a frontier model
alone already carries much of the outcome signal; its goal differs from ours, optimizing a
binary decision as a reviewer stand-in, whereas we validate a graded \emph{score} as a
measurement against two ground truths (tiers and ratings), bound the claim to low-end
triage, and characterize the score's reliability and failure modes rather than its
decision accuracy. A finding that LLM reviewers can overrate weak submissions
\citep{li2026llmreviewer} is then a reason to measure the low-end flag empirically, as H1
does, rather than assume it.

\paragraph{Contamination and temporal leakage.}
The central threat is that the grading model memorized the outcome in training, which
would inflate the relationship; LLM-generated review content is already widespread
\citep{liang2024monitoring}. We control for it by cohort choice: GPT-5.4's August~2025
cutoff \citep{openai2026gpt54} precedes the January~2026 release of ICLR~2026 decisions,
so the outcome could not have been in training. We add an arXiv-before-cutoff sensitivity split bounding residual
paper-text leakage (Appendix~\ref{app:robust}). A fully pre-cutoff contaminated contrast
and a prospective cohort, which would close leakage entirely, are deferred
(Appendix~\ref{sec:limits}); whether the model's \emph{priors} on authors and
institutions move the score is addressed by a separate prestige-perturbation experiment.

\paragraph{Limitations.}
The strongest threats follow; the full treatment is in Appendix~\ref{sec:limits}. The
reject class is ``weak \emph{relative to the venue bar},'' not weak absolutely, and the
decision is itself noisy \citep{cortes2021inconsistency,beygelzimer2023arbitrary}; we
mitigate by also validating against the mean rating, but no analysis exceeds its ground
truth, and that ground truth may itself be partly LLM-assisted
\citep{liang2024monitoring}, a ceiling on H4. The scale rests on the cheap-model bridge,
which is weakest at the low end, which is why we report the deployable flag on AIPR
(GPT-5.4) directly. The primary cohort is one venue in one field, so we do not claim
cross-field generalization. And the scorer is proprietary: the released data, labels, and
analysis code make the score-to-outcome study fully reproducible, but the
manuscript-to-score function is audited through its outputs rather than re-implemented, a
defensible posture for a deployed instrument that pre-registration before unblinding lets
a reader check rather than trust. We do not claim the score predicts acceptance, ranks
strong papers, or should drive any decision without a human in the loop.

\paragraph{What the human signal's structure implies.}
Individual reviewers are highly variable \citep{reviewbench2026}, which is why we validate
against the \emph{mean} rating: a score cannot agree with a ground truth more reliable
than that truth is with itself, part of why we lead with the low-end claim rather than
ranking the top.

\section{Conclusion}
\label{sec:conclusion}

We validated a training-free, prompt-only LLM manuscript score against the public
decisions and reviewer ratings of a major machine learning venue, under a
pre-registered, reproducible protocol. The score separates rejected from accepted
submissions (AUROC~$\aurocMini$), rises across decision tiers, and tracks reviewer
ratings ($\rho=\spearmanRatingMini$); it is strongest at the low end, where the lowest
quintile is rejected at \bottomRejectRateFrontier\% on the production model, with strong
work absent. Two findings locate the system's value. The validity comes mostly from the
model: a one-paragraph prompt on the same model already discriminates, indistinguishably
from the full pipeline ($\Delta$AUROC \aurocDiffFullNaive, \aurocDiffPrel). What the
engineering adds is reliability: AIPR's score barely moves across runs (\fullRunSD{}
vs.\ \naiveRunSD{} points) where the bare prompt swings, delivered with a grounded,
anchored review in one pass. Intelligence is not the bottleneck; reliability is. The
validated use is narrow: an automated first pass that surfaces weak work for human
attention, not a replacement for the judgment that follows.

\paragraph{Reproducibility.}
The scored table (keyed by public OpenReview submission identifiers), labels, analysis
code, and pre-registration are released
at \url{https://github.com/cogeor/aipr-score-validation}; one command regenerates every
figure and number. The full review text for identifiable
submissions is retained as a controlled audit artifact. The grading rubric, prompts, and
model configuration remain proprietary; the manuscript-to-score function is audited
through its outputs.

\paragraph{Competing interests.}
The author founded and operates AIPR (\url{https://aipr.pub}), the commercial system
evaluated in this study, and stands to benefit from a favorable result. The study was
self-funded; no external funding was received. The mitigations are structural rather
than rhetorical: the analysis plan was frozen at a public tag before any score was
joined to an outcome, every reported number regenerates from released code and data,
and the primary comparison includes a baseline designed to make the pipeline redundant.
Nothing in this study implies endorsement by ICLR, OpenReview, or the authors of any
cited submission.

\bibliographystyle{plainnat}
\bibliography{refs}

\appendix
\section{Pre-registration (verbatim)}
\label{app:prereg}
This section records the analysis plan as frozen before any AIPR score was
joined to any outcome. The primary hypothesis (H1), the primary metric
(bottom-quintile reject lift), the discrimination metric (AUROC), the trend test
(Jonckheere--Terpstra, Monte Carlo permutation), the bridge gate ($\rho\ge0.8$),
and the exclusion rules were all fixed in advance. The full plan is reproduced
verbatim in \texttt{DECISIONS.md} at the public tag
\texttt{prereg-iclr2026-v2} (\url{https://github.com/cogeor/aipr-score-validation});
the push/tag timestamp on that public remote is the freeze anchor (commit dates
are forgeable). We point to the tagged file rather than reprinting it here, so the
pre-registration a reader checks is the immutable public one and cannot drift from
a transcription.

\section{Grading configurations}
\label{app:configs}

\begin{figure}[h]
  \centering
  \includegraphics[width=\textwidth]{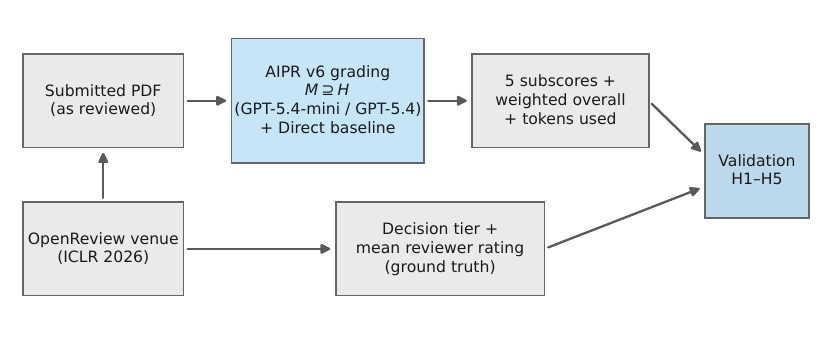}
  \caption{\textbf{Study design.} Each OpenReview submission supplies both the artifact
  reviewers saw (the submitted PDF) and the ground truth (decision tier and mean
  reviewer rating). AIPR grades the PDF with the full two-pass pipeline under two
  strictly nested configurations ($H\subseteq M$, differing only in the model:
  GPT-5.4-mini and GPT-5.4) plus the one-paragraph Direct baseline (V1); the resulting
  scores are validated against the ground truth (H1--H5). No model is tuned on the
  venue, decisions, or ratings. Interactive counterpart and live result figures:
  \url{https://aipr.pub/publications}.}
  \label{fig:design}
\end{figure}

\begin{table}[h]
  \centering
  \caption{Grading configurations. The two pipeline configurations are the cost
  ladder (cohort nesting $H\subseteq M$); they run the identical full two-pass
  pipeline and share the scoring formula (Eq.~\ref{eq:overall}), differing only in
  the reviewer model tier. The Direct baseline (Appendix~\ref{app:naive}) is graded on
  cohort $H$ so it pairs with a full-text score; it uses the same frontier model
  but a single prompt with no rubric or audit, and returns only an overall score.}
  \label{tab:configs-app}
  \begin{tabular}{lll}
    \toprule
    Configuration & Pipeline input / mode & Model tier \\
    \midrule
    AIPR (GPT-5.4-mini)  & full text, two-pass (+ audit)     & cheap \\
    AIPR (GPT-5.4)       & full text, two-pass (+ audit)     & frontier \\
    \midrule
    Direct (GPT-5.4) & full text, single one-paragraph prompt (no rubric/audit) & frontier \\
    \bottomrule
  \end{tabular}
\end{table}

\section{Data schema}
\label{app:schema}
The analysis consumes two tables. \texttt{submissions} (one row per submission):
identifier, venue, year, decision tier, tier rank, accept flag, mean reviewer
rating, rating SD, review count, sampling stratum, an exclusion flag with reason,
and optional manuscript metadata (primary area, page/word/token/reference counts).
\texttt{gradings} (one row per grading run): submission id, configuration, run
index, model name, pipeline version, the five dimension scores, the overall score,
the tokens used, and provenance ids. Contract invariants (tier/rank consistency,
uniform pipeline version, score ranges, referential integrity, cohort nesting
$H\subseteq M$) are asserted at load.

\section{Provenance and the grading boundary}
\label{app:provenance}
\paragraph{Decision-label derivation.} Each submission's raw OpenReview venue tag
is released verbatim in \texttt{decision\_raw} (e.g.\ \texttt{ICLR.cc/2026/Conference/Oral}).
The ordinal \texttt{decision\_tier} (reject/poster/oral) and \texttt{tier\_rank}
($0/1/2$) are derived from that tag by a deterministic map, so any reader can
re-derive every tier assignment from the released label; the load-time contract
asserts the tag is non-empty and that the tier, rank, and accept fields agree.

\paragraph{Inference-time access boundary.} During grading the reviewer model
received the submitted PDF bytes and nothing from the submission's OpenReview
forum: not the decision, the review text, the venue tag, or the reviewer ratings.
The grounded citation audit could query bibliographic records to check the
bibliography, but was not given access to OpenReview decision or review pages.
Combined with the grading model's documented knowledge cutoff
(31~August~2025 \citep{openai2026gpt54}, before the ICLR~2026 decisions and
reviews were released in January~2026; \S\ref{sec:discussion}), the outcome we
validate against could neither be read at grading time nor memorized in training.

\section{Deployment boundary, independence, and audit posture}
\label{app:posture}
This section collects the conflict-of-interest, deployment-ethics, and
audit-access commitments that the rest of the paper relies on but does not state
in one place.

\paragraph{Independence and conflict of interest.} This is an author-led
validation of a system the authors build, so the design is structured to be
checkable rather than taken on trust: the hypotheses, metrics, and exclusion
rules are pre-registered (Appendix~\ref{app:prereg}), the cohort is frozen before
unblinding, every figure and statistic is regenerated from released labels and
scores by the committed analysis code, and both error directions are reported in
full (\S\ref{sec:failure}). These mitigations narrow but do not remove the
incentive; an independent replication on a different venue (which the
author-built scorer cannot influence) is the strongest next step and is the
deferred second-venue study (\S\ref{sec:limits}, Appendix~\ref{app:robust}).

\paragraph{Deployment boundary: the validated action is prioritization, not
rejection.} The use we validate is routing a submission to human attention, not
deciding it. A low score is a flag that triggers a human look; a high score is
\emph{not} used for positive selection or to skip review (the failure-mode
asymmetry of \S\ref{sec:failure} is what licenses this). Any venue deployment
should keep a human accountable for every decision, log the model's outputs so a
flagged decision can be audited after the fact, and disclose to authors that
automated assistance was used. These are deployment conditions, not properties we
measured.

\paragraph{Proprietary-scorer audit posture.} The reproducibility paragraph
(\S\ref{sec:limits}) explains that the score-to-outcome analysis is fully open
while the manuscript-to-score function is closed. To keep the closed half
auditable rather than opaque, the scoring dimensions, the weighting rule
(Eq.~\ref{eq:overall}), the output schema (Appendix~\ref{app:schema}), the
pipeline version, and the model identifiers are reported in this paper, and the
exact prompt and configuration artifacts are retained under the frozen study tag
so they can be provided confidentially to editors or auditors on request. We do
not publish the prompts themselves, and we make no claim that a configuration
hash is published.

\paragraph{Baseline fairness.} The Direct one-paragraph prompt
(Appendix~\ref{app:naive}) is intentionally \emph{not} optimized: it runs the
same frontier model on the same submitted PDF with no rubric and no audit, and
its wording was frozen before unblinding. Its purpose is to isolate the
pipeline's contribution over the bare model, not to maximize baseline
performance, so we did not tune it post hoc to weaken or strengthen the
comparison.

\section{Design justification: statistical power and estimator validity}
\label{app:power}
The power and estimator checks below were run \emph{before} data collection and
do not use the real outcomes; they establish that the planned sample is adequate
and that the analysis code is correct. The one exception, marked as such, is the
post-hoc V1 minimum-detectable-effect analysis at the end of the power
paragraph, added after unblinding. All are reproduced by
\texttt{analysis/simulation.py}.

\begin{figure}[h]
  \centering
  \includegraphics[width=\textwidth]{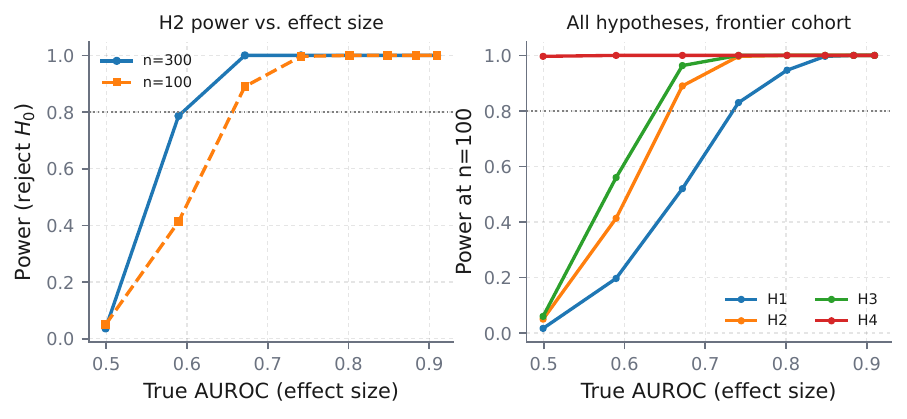}
  \caption{Pre-data power analysis. Left: power of the separation test (H2) vs.\
  true effect size at the two planned cohort sizes; dotted line is 80\% power.
  Right: power of all four hypotheses at the frontier cohort ($n=100$). Both
  generated by \texttt{simulation.py} on a synthetic generative model, not on
  study data.}
  \label{fig:power}
\end{figure}

\paragraph{Power.}
Under a generative model in which a latent paper quality drives the decision
tier, the reviewer rating, and (more noisily) the AIPR score, we swept the
effect-size knob and report power as a function of the \emph{true} AUROC rather
than a single guessed value (Fig.~\ref{fig:power}). At the planned cohort sizes
the design is well powered: for a moderate true effect the frontier cohort
($n=\NfullPrimary$) attains power $\powerHtwoN\%$ on the separation test (H2) and
$\powerHoneN\%$ on the primary low-end test (H1), and the cohort $M$
($n\approx\NminiPrimary$) is at ceiling. The minimum detectable effect for H2 at
$n=100$ and 80\% power is a true AUROC of about $\mdeAUROC$, well below any
plausible value, so a null result would be informative rather than a power
artifact. The value comparison (V1) is the post-hoc exception: under a paired
generative model whose two scores share a latent paper quality and whose
within-paper correlation is calibrated to the observed full-vs-direct score
correlation, the realized $n=\NfullPrimary$ design reaches 80\% power only for
a true paired AUROC gap of at least \mdeAUROCdiff. The observed gap
(\aurocDiffFullNaive) sits below that threshold, which is why
\S\ref{sec:res-value} reads V1 as criterion-not-met rather than as evidence of
equivalence. This analysis was added after unblinding and uses two observed
summary constants (the score-score correlation and the direct arm's AUROC),
never the outcome data.

\paragraph{Estimator validity.}
On data with a known population value, our 95\% BCa bootstrap confidence intervals
covered the truth $\bootCoverage\%$ of the time (nominal 95\%; the simpler
percentile interval under-covers at this sample size, which is why we use BCa). Under a true null
(no trend across tiers), the Jonckheere--Terpstra permutation test rejected at
$\jtTypeOne\%$ (nominal 5\%) and its $p$-values were uniform (Kolmogorov--Smirnov
$p=\jtUniformP$). The analysis therefore controls error at its stated levels.
Additionally, a label-shuffle null control is asserted at analysis time: with
decision labels permuted, the overall-score AUROC is $\approx 0.5$.

\section{Robustness and secondary analyses}
\label{app:robust}

\begin{figure}[h]
  \centering
  \includegraphics[width=\textwidth]{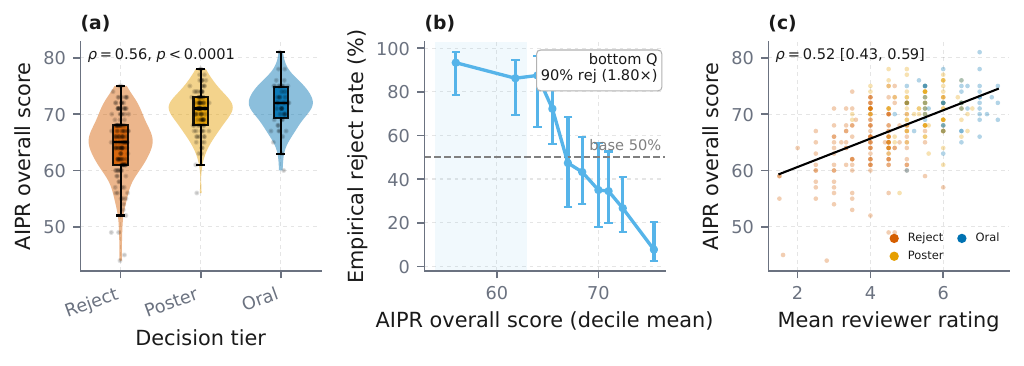}
  \caption{\textbf{The AIPR score agrees with the human outcome three ways}
  (cohort $M$; the detailed companion to the main-text validity result of
  \S\ref{sec:res-validity}). \textbf{(a)} Overall score across the ordered decision
  tiers (violin, box, jittered points), rising monotonically (H3). \textbf{(b)}
  Reject rate by score decile with Wilson 95\% intervals; shaded band is the bottom
  score quintile, dashed line the base reject rate, so low scores reliably mark
  rejected work (H1). \textbf{(c)} Score vs.\ mean reviewer rating, coloured by tier,
  with a least-squares trend and Spearman $\rho$ (H4). The AUROC scalar (H2) is in
  Table~\ref{tab:headline}; the overlaid ROC curves are the main-text
  Fig.~\ref{fig:naive}a. Interactive counterpart (per-point subscores on hover):
  \url{https://aipr.pub/publications\#fig-validation}.}
  \label{fig:validation}
\end{figure}

\begin{figure}[h]
  \centering
  \includegraphics[width=\textwidth]{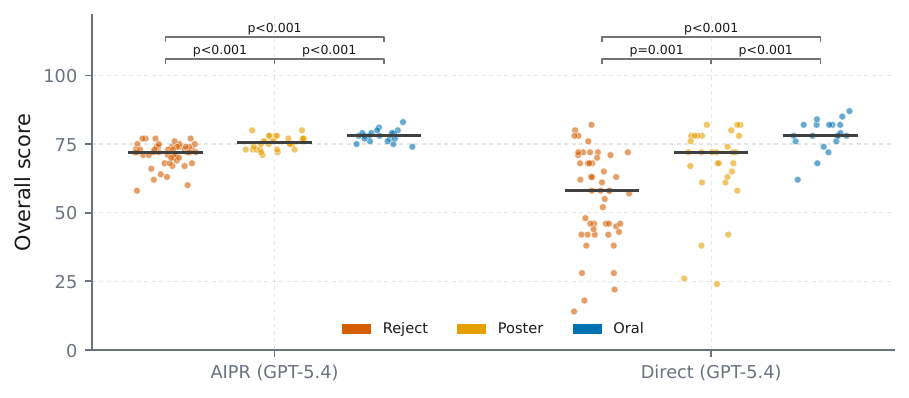}
  \caption{Overall-score distribution across the three ordered decision tiers, the
  AIPR (GPT-5.4) vs.\ the Direct (GPT-5.4) baseline (cohort $H$), as
  jittered points with a per-tier median bar. Brackets give pairwise two-sided
  Mann--Whitney $p$-values. Both graders separate the tiers; this is the same
  strong-baseline result as the overlaid ROC (Fig.~\ref{fig:naive}a), seen
  as per-tier distributions. The paired AUROC difference is not statistically
  resolved, so this figure should not be read as an equivalence result. The three-grader ROC
  curves themselves, with AUROC and 95\% CIs, are the main-text
  Fig.~\ref{fig:naive}a.}
  \label{fig:score-dist}
\end{figure}

\begin{figure}[h]
  \centering
  \includegraphics[width=0.75\columnwidth]{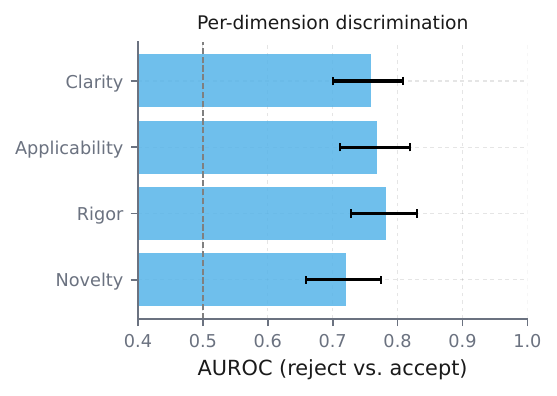}
  \caption{Per-dimension reject-vs-accept AUROC (cohort $M$) with 95\% CIs.
  Rigor and applicability carry the most discriminative signal; novelty the
  least.}
  \label{fig:subscore-auroc}
\end{figure}

\begin{table}[h]
  \centering
  \caption{Per-dimension AUROC (cohort $M$).}
  \label{tab:subscores-app}
  \begin{tabular}{lcc}
\toprule
Dimension & AUROC (reject vs.\ accept) & BH $q$ \\
\midrule
Novelty & 0.72 [0.66, 0.77] & 0.000 \\
Rigor & 0.78 [0.73, 0.83] & 0.000 \\
Applicability & 0.77 [0.71, 0.82] & 0.000 \\
Clarity & 0.76 [0.70, 0.81] & 0.000 \\
\bottomrule
\end{tabular}

\end{table}

\begin{table}[h]
  \centering
  \caption{Pairwise Pearson correlation among the four informative subscores
  (cohort M, $n=\haloN$; citation excluded: pinned at 100, zero variance). The
  dimensions share a general-quality component (mean off-diagonal
  $r=\haloMeanR$) but retain distinct variance: the weakest pair
  (novelty--clarity, $r=\haloMinR$) is the most distinct, the strongest
  (rigor--clarity, $r=\haloMaxR$) is the expected writing-quality confound. Not a
  single collapsed factor, but not orthogonal either; see
  \S\ref{sec:discussion}.}
  \label{tab:subscore-corr}
  \begin{tabular}{lcccc}
\toprule
 & Novelty & Rigor & Applicability & Clarity \\
\midrule
Novelty & 1 & 0.49 & 0.32 & 0.30 \\
Rigor & 0.49 & 1 & 0.61 & 0.73 \\
Applicability & 0.32 & 0.61 & 1 & 0.50 \\
Clarity & 0.30 & 0.73 & 0.50 & 1 \\
\bottomrule
\end{tabular}

\end{table}

\begin{figure}[h]
  \centering
  \includegraphics[width=0.6\columnwidth]{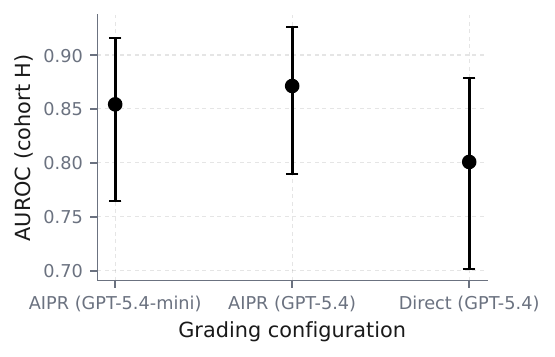}
  \caption{AUROC for AIPR (GPT-5.4-mini), AIPR (GPT-5.4), and the Direct (GPT-5.4)
  one-paragraph baseline on the shared cohort $H$, showing whether the frontier model
  adds discriminative signal over the cheap model and over the bare prompt (both AIPR
  configs run the full two-pass pipeline).}
  \label{fig:nested-auroc}
\end{figure}

\begin{figure}[h]
  \centering
  \includegraphics[width=\textwidth]{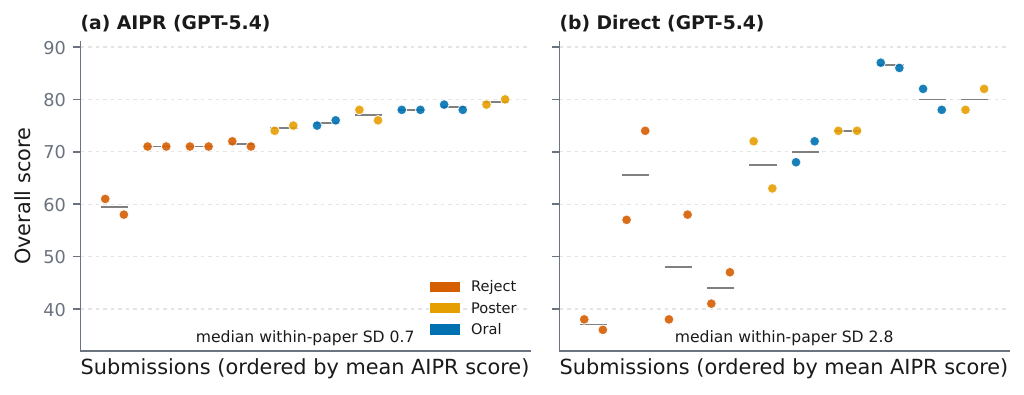}
  \caption{Run-to-run scoring spread, one point per re-grading. Each column is a
  variance-sub-study paper; its repeated re-gradings stack vertically and are
  coloured by the paper's human decision tier. Both panels share one $x$-order
  (ascending mean AIPR score), so AIPR's tight columns (a) read directly against
  the baseline's wider ones (b); the median within-paper SD annotated on each
  panel (\fullRunSD{} vs.\ \naiveRunSD{} points) resolved to the individual
  grading. Runs $0$ are excluded for both graders (\S\ref{sec:res-value}): run $0$
  of the frontier config is the original cohort's citation-pinned config-state, not
  stochastic noise.}
  \label{fig:run-variance}
\end{figure}

\paragraph{Multiple comparisons.} The five per-dimension discrimination tests are
controlled at FDR $0.05$ by the Benjamini--Hochberg procedure; adjusted $q$-values
accompany Table~\ref{tab:subscores-app}. The confirmatory hypotheses H1--H4 are
pre-specified single tests and are not part of this family.

\begin{figure}[h]
  \centering
  \includegraphics[width=0.6\columnwidth]{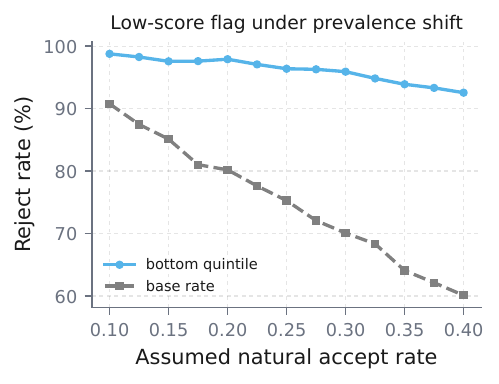}
  \caption{Low-score flag under prevalence shift. AUROC is prevalence-invariant,
  but a flag's precision is not, so we re-weight the balanced cohort across a range
  of assumed natural acceptance rates and show the lowest-score band stays well
  above the base reject rate throughout.}
  \label{fig:supp-cond}
\end{figure}

\paragraph{Prevalence re-weighting.} Because cohort $M$ over-samples the rarer
accept tiers relative to the venue's natural prevalence,
we re-weight to a range of natural acceptance rates (Fig.~\ref{fig:supp-cond});
the discrimination (AUROC) is prevalence-invariant and the low-score
flag's precision remains above base rate across the plausible range. At the
venue's actual prevalence (\natAcceptRate\% accept, base reject rate
\natBaseReject\%), the bottom-quintile reject precision is \natBottomPrecision\%.

\paragraph{Second-venue replication (deferred).} A second-venue replication
(ICLR~2025, AIPR (GPT-5.4-mini), as a fully pre-cutoff contaminated contrast) is reserved
for future work; this first analysis is ICLR~2026 only. The analysis pipeline
emits this section automatically once a 2025 export is present.

\begin{figure}[h]
  \centering
  \includegraphics[width=\textwidth]{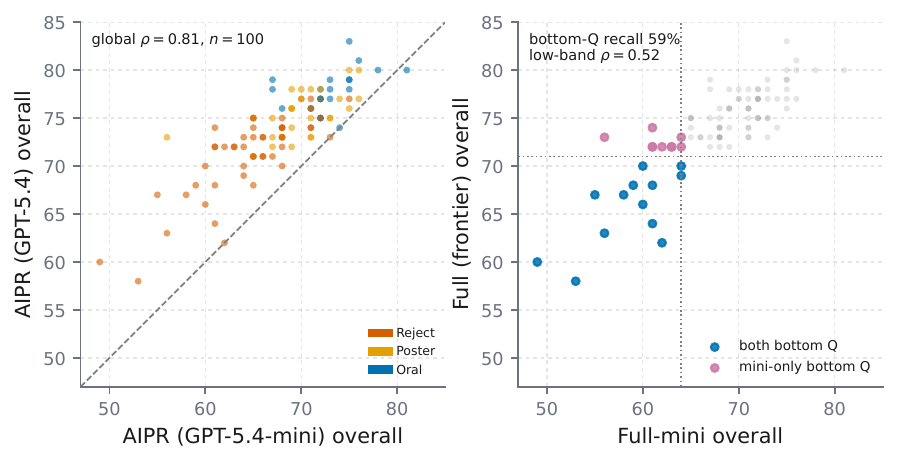}
  \caption{Mini-to-frontier bridge on the paired cohort $H$. Left: AIPR (GPT-5.4-mini) vs.\ AIPR (GPT-5.4)
  overall score, coloured by the human decision tier, dashed line identity (global
  agreement). Right: bottom-quintile
  membership (papers in both scores' bottom quintile vs.\ mini-only); dotted
  lines are the per-score quintile thresholds. The low-end recall and within-band
  $\rho$ quantify the bottom-quintile agreement, which is weaker than the global
  correlation (H5); the deployable low-end flag is therefore also computed on the
  production score directly (\S\ref{sec:res-validity}).
  Interactive counterpart (per-dimension subscores on hover):
  \url{https://aipr.pub/publications\#fig-bridge}.}
  \label{fig:bridge}
\end{figure}

\paragraph{Mini-to-frontier bridge (H5).} On the paired cohort $H$, the cheap-model
AIPR (GPT-5.4-mini) score and the AIPR (GPT-5.4) score agree closely
(Fig.~\ref{fig:bridge}, left; $\rho =$~\bridgeRhoFull, $n=\bridgeN$), clearing the
pre-registered $\rho\ge0.8$ gate that licenses the large-$N$ cohort-$M$ results as a
proxy for the production score (\S\ref{sec:results}). Because a global
correlation can mask low-end disagreement, we also report agreement restricted to
the cohort-$M$ bottom quintile, $\rho =$~\bridgeLowRhoFull, with \bridgeOverlap\% of the
cohort-$M$ bottom quintile also in the frontier bottom quintile
(Fig.~\ref{fig:bridge}, right). The bridge is strong globally but weaker in the
bottom quintile, where the deployable claim lives (\S\ref{sec:limits}); this is
why the low-end flag (H1) is also computed on the production score directly
(\S\ref{sec:res-validity}), so the deployable claim does not route through the
weak low-end bridge.

\begin{figure}[h]
  \centering
  \includegraphics[width=0.6\columnwidth]{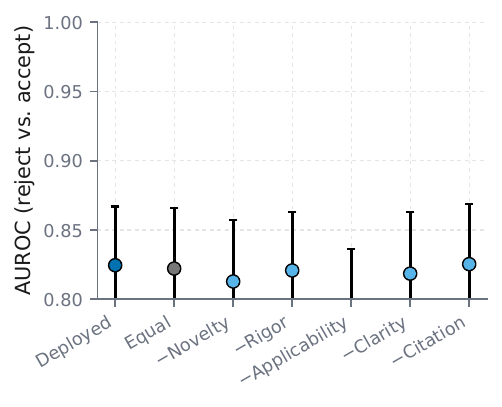}
  \caption{Reject-vs-accept AUROC (cohort $M$) under the deployed weighting,
  equal weighting, and each leave-one-dimension-out weighting, with 95\% CIs. The
  result is stable across weightings.}
  \label{fig:weightrobust}
\end{figure}

\paragraph{Weighting robustness.} The deployed overall weights five dimensions
(Eq.~\ref{eq:overall}) on substantive grounds, frozen before this study. To show
the headline does not hinge on the exact (proprietary) coefficients, we recompute
the overall under equal weights and under each leave-one-dimension-out weighting
and report discrimination plus rank agreement with the deployed score
(Table~\ref{tab:weightrobust}, Fig.~\ref{fig:weightrobust}). The equal-weight
score reaches AUROC~\aurocEqualWeightFull and ranks papers nearly identically to
the deployed score ($\rho=\rhoEqualWeight$); every leave-one-out variant stays in
the narrow band AUROC $\in[\looAurocMin,\looAurocMax]$. These are robustness
checks, not a search for a better weighting.

\begin{table}[h]
  \centering
  \caption{Headline discrimination under the deployed, equal, and
  leave-one-dimension-out weightings (cohort $M$), with rank agreement against
  the deployed score.}
  \label{tab:weightrobust}
  \begin{tabular}{lcc}
\toprule
Weighting & AUROC (reject vs.\ accept) & $\rho$ vs.\ deployed \\
\midrule
Deployed weights & 0.82 [0.78, 0.87] & --- \\
Equal weights & 0.82 [0.77, 0.87] & 0.92 \\
\midrule
Drop novelty & 0.81 [0.76, 0.86] & 0.89 \\
Drop rigor & 0.82 [0.77, 0.86] & 0.99 \\
Drop applicability & 0.79 [0.74, 0.84] & 0.88 \\
Drop clarity & 0.82 [0.77, 0.86] & 0.99 \\
Drop citation & 0.83 [0.78, 0.87] & 0.99 \\
\bottomrule
\end{tabular}

\end{table}

\begin{figure}[h]
  \centering
  \includegraphics[width=0.6\columnwidth]{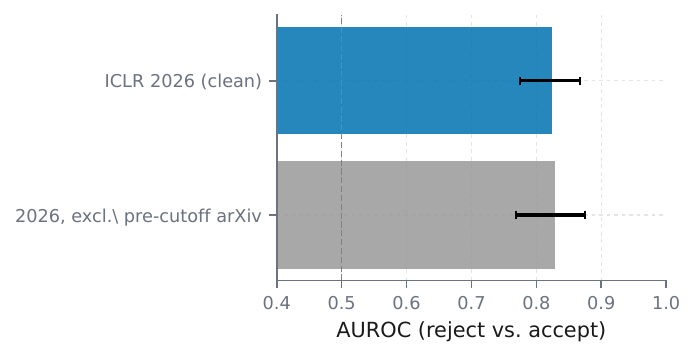}
  \caption{Reject-vs-accept AUROC for the clean primary cohort and the primary
  cohort excluding pre-cutoff arXiv papers, with 95\% CIs. The headline is
  unchanged by the arXiv exclusion: paper-text leakage is not driving the result.
  The fully pre-cutoff ICLR~2025 contaminated contrast is deferred to future work.}
  \label{fig:contamination}
\end{figure}

\paragraph{Temporal leakage controls.} The primary cohort's decisions postdate
the grading model's knowledge cutoff, so the outcome cannot have been memorized.
We bound the residual paper-text risk by excluding the \nArxivPrior{} submissions
whose arXiv preprint predates the cutoff: this leaves the headline essentially
unchanged (AUROC~\aurocNoArxivPriorFull), so the model reading the manuscript (as a
reviewer would) is not itself a confound (Table~\ref{tab:contam},
Fig.~\ref{fig:contamination}). A fully pre-cutoff contaminated contrast
(ICLR~2025, AIPR (GPT-5.4-mini)) is deferred to future work; the prospective check (grading a
future venue before decisions are released) remains the only complete control and
is reported as follow-up. Whether the model's priors on authors and institutions
move the score is a separate question, addressed by the prestige-perturbation
experiment rather than by a grading configuration.

\begin{table}[h]
  \centering
  \caption{Temporal leakage controls (cohort $M$): the clean primary cohort
  and the primary cohort with pre-cutoff arXiv papers excluded. The fully
  pre-cutoff ICLR~2025 contaminated contrast is deferred to future work.}
  \label{tab:contam}
  \begin{tabular}{lcc}
\toprule
Cohort & $n$ & AUROC (reject vs.\ accept) \\
\midrule
ICLR 2026 (clean) & 300 & 0.82 [0.78, 0.87] \\
2026, excl.\ pre-cutoff arXiv & 230 & 0.83 [0.77, 0.87] \\
\bottomrule
\end{tabular}

\end{table}

\begin{figure}[h]
  \centering
  \includegraphics[width=0.62\columnwidth]{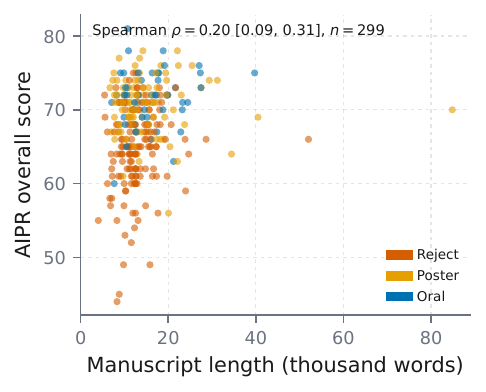}
  \caption{AIPR overall score against manuscript length (cohort $M$),
  coloured by the human decision tier. If the score rewarded length it would climb
  with word count; instead the rank correlation is weak ($\rho=\lenRhoWord$) and the
  decision tiers overlap across the length axis. Support check, not a headline
  result.}
  \label{fig:covariate}
\end{figure}

\paragraph{Manuscript-length confounding.} A score that simply rewarded longer or
more polished manuscripts would be confounded with length rather than quality. We
correlate the AIPR overall with manuscript page count, word count, reference
count, and figure count on the cohort $M$ (Table~\ref{tab:lengthconfound}). The
rank correlations are weak ($\rho=\lenRhoPage$ with page count,
$\rho=\lenRhoWord$ with word count, $\rho=\lenRhoRefs$ with reference count), so
the discrimination is not explained by manuscript length.
Figure~\ref{fig:covariate} shows the word-count relationship directly: the tiers
overlap across the length axis rather than sorting along it.

\begin{table}[h]
  \centering
  \caption{Rank correlation of the AIPR overall with manuscript-length metrics
  (cohort $M$). Weak throughout: the score is not a length proxy.}
  \label{tab:lengthconfound}
  \begin{tabular}{lc}
\toprule
Manuscript metric & Spearman $\rho$ with AIPR overall \\
\midrule
Page count & 0.24 [0.13, 0.35] \\
Word count & 0.20 [0.09, 0.31] \\
Reference count & -0.09 [-0.20, 0.03] \\
Figure count & 0.11 [-0.00, 0.23] \\
\bottomrule
\end{tabular}

\end{table}

\paragraph{Covariate-control discrimination (descriptive).} A reviewer asked
whether the score-outcome relationship is merely a proxy for manuscript surface
features or subject area. We audit this descriptively, as an exploratory
confound check, not a replacement for the pre-registered AUROC, so we read it for
direction rather than as a confirmatory test. We fit a logistic model of the
accept outcome on the AIPR overall together with primary area and six
manuscript-surface covariates (page, word, reference, and figure counts, rating
SD, and review count), and compare its cross-validated AUROC against a score-only
model under the same stratified five-fold protocol (Table~\ref{tab:covcontrol}).
Adding the covariates does not collapse the discrimination: AIPR (GPT-5.4-mini) moves from
\scoreAucMini{} (score only) to \covAucMini{} ($n=\nCovMini$) and the frontier
cohort from \scoreAucFrontier{} to \covAucFrontier{} ($n=\nCovFrontier$), so no
single area or covariate set explains away the score-outcome relationship; the
covariates carry incremental signal but the score is not a stand-in for them.

\begin{table}[h]
  \centering
  \caption{Covariate-control cross-validated AUROC (descriptive): the AIPR
  overall alone vs.\ the overall plus primary area and manuscript-surface
  covariates, under one stratified five-fold protocol per cohort. Exploratory
  confound audit, not a confirmatory replacement for the headline AUROC.}
  \label{tab:covcontrol}
  \begin{tabular}{lccc}
\toprule
Cohort & Score-only AUROC & $+$covariates AUROC & $n$ \\
\midrule
AIPR (GPT-5.4-mini) & 0.83 & 0.87 & 299 \\
AIPR (GPT-5.4) & 0.87 & 0.83 & 100 \\
\bottomrule
\end{tabular}

\end{table}

\paragraph{Within-tier score--rating agreement (descriptive).} The bounded claim
is that the score is a low-end triage signal, not a fine ranking of strong
papers. As a descriptive check we compute the Spearman correlation between the
AIPR overall and the mean reviewer rating \emph{within} each decision subgroup on
the cohort $M$ (Table~\ref{tab:withintier}). Within the reject tier the
correlation is the strongest ($\rho=\rhoWithinReject$); within the accepted
subgroup (poster $+$ oral) it is weaker ($\rho=\rhoWithinAccepted$, with
$\rho=\rhoWithinPoster$ poster-only and $\rho=\rhoWithinOral$ oral-only).
The within-accepted correlation is weaker, consistent with the bounded low-end
claim: the score separates clearly weak work from the rest but does not finely
rank papers that have already cleared the bar.

\begin{table}[h]
  \centering
  \caption{Within-subgroup Spearman correlation of the AIPR overall with the mean
  reviewer rating (cohort $M$, descriptive). The within-accepted
  correlation is weaker, consistent with a low-end triage signal rather than a
  fine ranking of strong papers.}
  \label{tab:withintier}
  \begin{tabular}{lcc}
\toprule
Subgroup & $n$ & Spearman $\rho$ (overall vs.\ rating) \\
\midrule
Reject & 150 & 0.29 \\
Poster & 100 & 0.03 \\
Oral & 50 & 0.10 \\
Accepted (poster $+$ oral) & 150 & 0.11 \\
\bottomrule
\end{tabular}

\end{table}

\paragraph{Bottom-band tie/threshold sensitivity (descriptive).} The AIPR overall
is an integer score, so the exact membership of the bottom score quintile is
mildly tie-dependent (the reference quintile is uneven, $n=59$, because of ties at
the cut). A reviewer asked whether the low-score flag survives the choice of
bottom-band rule. We compute the bottom-band reject rate, the lift over the base
reject rate, and the oral rate under several definitions: the strict bottom
quintile (\mbox{$\text{score}<\text{quantile}(0.2)$}, the pre-registered band-0
membership), a deterministic bottom-$K$ ($K=60$ after sorting by score then
submission order, so ties never resolve by chance), and three fixed cutoffs
(\mbox{$\text{score}\le63$}, $64$, $65$) (Table~\ref{tab:bandsens}). Across all
\bandSensNdefs{} definitions the reject rate stays at least \bandSensMinReject\%
and the lift over base at least \bandSensMinLift{}, so the flag is robust to the
tie/threshold rule rather than an artifact of the quintile cut.

\begin{table}[h]
  \centering
  \caption{Bottom-band tie/threshold sensitivity of the low-score flag
  (AIPR (GPT-5.4-mini), descriptive). Per band definition: reject rate, lift over the base
  reject rate, and oral rate. The flag holds across the quintile, deterministic
  bottom-$K$, and fixed-cutoff rules.}
  \label{tab:bandsens}
  \begin{tabular}{lcccc}
\toprule
Bottom-band definition & $n$ & Reject rate & Lift & Oral rate \\
\midrule
strict quintile & 59 & 90\% & 1.80 & 3\% \\
bottom-60 & 60 & 90\% & 1.80 & 3\% \\
score<=63 & 59 & 90\% & 1.80 & 3\% \\
score<=64 & 75 & 89\% & 1.79 & 3\% \\
score<=65 & 94 & 87\% & 1.74 & 3\% \\
\bottomrule
\end{tabular}

\end{table}

\paragraph{Reviewer-disagreement moderation (descriptive).} Does the score track
the outcome worse where the human reviewers disagree? We read this two ways on
both cohorts (Table~\ref{tab:disagree}). First, the Spearman correlation between
the magnitude of the score-vs-rating rank residual and the reviewer rating SD is
weak ($\rho=\disagreeRhoMini$ cohort $M$, $\rho=\disagreeRhoFrontier$ frontier):
the score's disagreement with the human ranking does not grow strongly with
reviewer disagreement. Second, splitting each cohort at the median rating SD, the
AUROC on the high-disagreement half (\disagreeAucHighMini{} cohort $M$,
\disagreeAucHighFrontier{} frontier) is close to the low-disagreement half
(\disagreeAucLowMini{} and \disagreeAucLowFrontier{}). We treat this
descriptively: the ground truth is itself noisier where reviewers disagree, so a
modest AUROC difference is expected and is not read as a model failure. The signal
is not concentrated in the easy, high-agreement cases.

\begin{table}[h]
  \centering
  \caption{Reviewer-disagreement moderation (descriptive, both cohorts). The
  residual--rating-SD Spearman and the AUROC of the AIPR overall split at the
  median reviewer rating SD. Disagreement does not strongly moderate the
  score--outcome relationship.}
  \label{tab:disagree}
  \begin{tabular}{lcccc}
\toprule
Cohort & Residual--SD $\rho$ & AUROC (low disagr.) & AUROC (high disagr.) & $n$ low/high \\
\midrule
AIPR (GPT-5.4-mini) & 0.09 & 0.86 & 0.79 & 161/139 \\
AIPR (GPT-5.4) & 0.11 & 0.91 & 0.84 & 53/47 \\
\bottomrule
\end{tabular}

\end{table}

\paragraph{Area/subfield subgroup audit (descriptive).} Is the result
concentrated in a single ICLR primary area? We audit the cohort $M$ per
\texttt{primary\_area}, pooling areas with fewer than eight submissions into a
single \emph{other} row so tiny cells are not over-interpreted; for each area we
report the accept rate, mean AIPR score, the score--rating Spearman, and the
AUROC where both decision classes are present (Table~\ref{tab:areasub}). Across
the \nAreaRows{} reported rows the per-area AUROC ranges from \areaAurocMin{} to
\areaAurocMax{}. These per-area estimates are noisy and are treated as
descriptive; no single area accounts for the headline relationship, which is
present broadly rather than driven by one subfield.

\begin{table}[h]
  \centering
  \caption{Per-area subgroup audit (AIPR (GPT-5.4-mini), descriptive). Accept rate, mean
  AIPR score, score--rating Spearman, and AUROC per primary area; areas with
  $n<8$ are pooled into \emph{other}. Cells are small and noisy; read for breadth,
  not point estimates.}
  \label{tab:areasub}
  {\small
\begin{tabular}{p{2.4in}ccccc}
\toprule
Primary area & $n$ & Accept rate & Mean score & $\rho$ (score--rating) & AUROC \\
\midrule
foundation or frontier models, including LLMs & 41 & 63\% & 67.4 & 0.67 & 0.86 \\
applications to computer vision, audio, language, and other modalities & 38 & 58\% & 68.5 & 0.50 & 0.85 \\
datasets and benchmarks & 30 & 67\% & 68.0 & 0.57 & 0.97 \\
generative models & 25 & 44\% & 70.0 & 0.52 & 0.84 \\
alignment, fairness, safety, privacy, and societal considerations & 24 & 62\% & 66.2 & 0.68 & 0.88 \\
reinforcement learning & 20 & 35\% & 66.2 & 0.48 & 0.71 \\
other topics in machine learning (i.e., none of the above) & 17 & 18\% & 66.8 & 0.32 & 0.77 \\
optimization & 14 & 29\% & 68.4 & 0.34 & 0.88 \\
transfer learning, meta learning, and lifelong learning & 14 & 29\% & 68.8 & 0.62 & 0.96 \\
applications to physical sciences (physics, chemistry, biology, etc.) & 13 & 69\% & 68.8 & 0.23 & 0.75 \\
interpretability and explainable AI & 12 & 50\% & 63.1 & 0.81 & 1.00 \\
unsupervised, self-supervised, semi-supervised, and supervised representation learning & 12 & 50\% & 69.4 & 0.66 & 0.92 \\
learning on time series and dynamical systems & 8 & 25\% & 67.5 & 0.67 & 1.00 \\
other & 32 & 47\% & 67.8 & 0.55 & 0.89 \\
\bottomrule
\end{tabular}}

\end{table}

\begin{figure}[h]
  \centering
  \includegraphics[width=0.6\columnwidth]{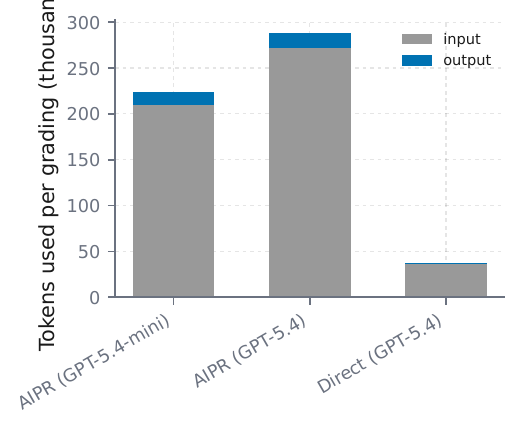}
  \caption{Mean tokens used per grading configuration (primary venue), input vs.\
  output. Both pipeline configurations run the two-pass pipeline (reviewer +
  audit), differing only in the model tier; the Direct baseline is a single prompt on
  the frontier model.}
  \label{fig:cost}
\end{figure}

\paragraph{Grading cost.} The cost design (\S\ref{sec:methods}) rests on the
AIPR (GPT-5.4-mini) configuration being far cheaper than the AIPR (GPT-5.4)
configuration. Mean tokens used per grading (Table~\ref{tab:cost},
Fig.~\ref{fig:cost}) are \costFullMiniTok{} for AIPR (GPT-5.4-mini) and \costFullTok{} for
full: the frontier configuration uses \costFullVsMini$\times$ the tokens of
AIPR (GPT-5.4-mini) (the same two-pass pipeline, but the costlier model tier). This is why
the relationship is established on the large cohort $M$ and only confirmed on
the budgeted frontier cohort.

\begin{table}[h]
  \centering
  \caption{Mean tokens used per grading configuration (primary venue). Input =
  prompt/manuscript tokens read; output = generated tokens.}
  \label{tab:cost}
  \begin{tabular}{lcccc}
\toprule
Configuration & Input & Output & Total & $n$ runs \\
\midrule
AIPR (GPT-5.4-mini) & 209.3k & 14.4k & 223.8k & 300 \\
AIPR (GPT-5.4) & 271.9k & 16.0k & 287.9k & 120 \\
Direct (GPT-5.4) & 36.8k & 0.7k & 37.4k & 120 \\
\bottomrule
\end{tabular}

\end{table}

\paragraph{Population boundary: every eligible submission accounted for.} The
pre-registered exclusions (\S\ref{sec:data}; DECISIONS.md~\S4) commit to a
ledger in which every eligibility-screened submission is either in-population
(eligible to be graded) or excluded with a stated reason, never silently
dropped. We report that ledger here (Table~\ref{tab:popboundary}):
13{,}718 submissions are in-population (passed eligibility), from which
the stratified graded cohort M ($n=300$) is sampled, and 6{,}096
screened-out submissions are excluded before any primary metric, for 19{,}814
eligibility-screened submissions in total. The excluded set is dominated by
submissions with no reviewer signal to validate against: 5{,}188
(85.1\% of the excluded set) are withdrawn before review and
908 (14.9\%) are desk-rejected. This ledger
reports counts and reasons only: the excluded submissions are accounted for by
disposition, but the manuscript covariates used in the confound checks above were
not extracted for ungraded submissions, so a covariate-level comparison of the
sampled cohort to the full eligible population is not available
(\S\ref{sec:limits}).

\begin{table}[h]
  \centering
  \caption{Population boundary: the in-population eligible set (and the graded
  cohort M sampled from it) beside the eligible-but-excluded ledger, broken down
  by exclusion reason. Every eligibility-screened submission is accounted for by
  disposition (DECISIONS.md~\S4); the ledger carries counts and reasons only, not
  manuscript covariates.}
  \label{tab:popboundary}
  \begin{tabular}{lrr}
\toprule
Disposition & Submissions & Share of excluded \\
\midrule
In-population (eligible, not excluded) & 13718 & --- \\
\quad of which graded (cohort M sample) & 300 & --- \\
\midrule
\textbf{Eligible-but-excluded} & \textbf{6096} & \textbf{100.0\%} \\
\quad Withdrawn before review & 5188 & 85.1\% \\
\quad Desk-rejected & 908 & 14.9\% \\
\midrule
\textbf{Total eligibility-screened} & \textbf{19814} & --- \\
\bottomrule
\end{tabular}

\end{table}

\paragraph{Per-review rating sensitivities (available on request).} The released
two-CSV contract carries the \emph{mean} reviewer rating per submission, not the
individual per-review ratings. Sensitivities that require the per-review
distribution, namely a human-review reliability ceiling (inter-reviewer agreement
as an upper bound on any score--rating correlation) and alternate rating
aggregations (median or trimmed mean in place of the mean), are therefore not
computed here.
We flag them as future work rather than report them: synthesizing per-review
ratings to populate these checks would be a fabrication, so they await a release
that includes the per-review ratings and are available on request.

\section{Direct baseline (V1): full detail}
\label{app:naive}
This appendix is the detailed companion to the value comparison of
\S\ref{sec:res-value} (the pre-registered experiment V1): the verbatim prompt, the
matched-accept-rate operating points, and the full table. The experiment
separates the contribution of the AIPR pipeline from the contribution of the
underlying model: we grade every frontier-cohort submission a second way, in which
the same model reads the same submitted PDF, but in place of the two-pass
rubric-and-audit pipeline it receives a single one-paragraph prompt and returns
one overall score with an accept/reject call. We publish the prompt verbatim, so
the baseline is realistic rather than a strawman or a covertly optimized variant:

\begin{quote}
\emph{I'm reviewing this paper for ICLR, a top machine learning conference. Read
it and tell me how good it is: give a single overall quality score from 0 to 100
(where higher means more likely to deserve acceptance) and a one or two sentence
reason for the score.}
\end{quote}

Because Direct (GPT-5.4) produces only an overall score, the comparison is on
score and decision alone (Table~\ref{tab:naive}, Fig.~\ref{fig:naive}):
discrimination (AUROC), accept/reject agreement at the pre-registered AIPR@60
operating point (predict accept iff overall $\ge 60$), and run-to-run reliability
(median within-paper SD over repeated runs). To be sure the comparison is not an
artifact of forcing the baseline onto AIPR's cutoff, we also score each method at
its own threshold matched to the human accept rate. On discrimination the AIPR
pipeline is higher but the paired difference is not statistically significant
(\aurocFullFull{} against \aurocNaiveFull{}; $\Delta$AUROC
\aurocDiffFullNaive, 95\% CI \aurocDiffCI, \aurocDiffPrel). This is not an
equivalence test; the paired cohort cannot rule out a small AIPR advantage or a
small baseline advantage over an already strong baseline. The fixed AIPR@60 cutoff is a pre-registered diagnostic rather than
evidence of decision-level agreement: the production score is compressed high and anchored
near the venue bar, so a 60 threshold predicts almost every paper accept and is not
venue-calibrated (\opSixtyFullAcc\% against \opSixtyNaiveAcc\% balanced accuracy for
AIPR and Direct (GPT-5.4)); we therefore rest the decision comparison on threshold-free
AUROC and report the fixed cutoff only for completeness. Where the pipeline separates
from the bare prompt is reliability: median within-paper SD \fullRunSD{} against
\naiveRunSD{} points. That reliability, with the grounded, anchored review produced
in the same pass, is what the engineered pipeline adds over the same model used
directly, not a higher score.

\begin{table}[h]
  \centering
  \caption{Direct (GPT-5.4) vs.\ AIPR (GPT-5.4) on the frontier
  cohort (same model, same PDF): discrimination (AUROC), accept/reject agreement
  at AIPR@60 (balanced accuracy), and run-to-run reliability (median within-paper
  SD).}
  \label{tab:naive}
  \begin{tabular}{lcc}
\toprule
Metric & AIPR (GPT-5.4) & Direct (GPT-5.4) \\
\midrule
AUROC (reject vs.\ accept) & 0.87 [0.79, 0.93] & 0.80 [0.70, 0.88] \\
Balanced accuracy @ 60 & 51.0\% & 72.0\% \\
Median run-to-run SD & 0.7 & 2.8 \\
\bottomrule
\end{tabular}

\end{table}

\section{Failure-mode case studies}
\label{app:cases}
Table~\ref{tab:casestudy} gives the matched examples behind \S\ref{sec:failure}:
AIPR's verbatim weakness beside a verbatim excerpt from the human reviews of the
\emph{same} frontier-cohort submission. \ifanonymous Submissions are described by
outcome and AIPR overall rather than named, as a courtesy to their authors.\else
Most rows are described by outcome and AIPR overall rather than named, as a
courtesy to their authors; the two exemplars named in the main text
(\S\ref{sec:results}, \S\ref{sec:failure}) --- the top-scoring oral and the
highest-AIPR-scored reject --- are public ICLR~2026 submissions, cited there for
concreteness. Submissions the score places low are never named.\fi{} The review excerpts are trimmed for length with
ellipses; the untrimmed text lives in the unreleased study artifact. The top block is correctly-flagged rejects; the lower two
are the error classes.

\begin{table}[h]
  \centering
  \footnotesize
  \caption{Matched case studies on the frontier cohort: AIPR's flagged weakness
  beside a verbatim human-review excerpt for the same submission. Top: weak work the
  score caught: the model and the reviewers name the same defect. Bottom: the two
  error classes, a rejected paper the score rated highly and accepted (oral) papers
  it underrated. Human-review excerpts are verbatim from the official anonymous
  ICLR~2026 reviews on OpenReview (CC~BY~4.0); excerpts are shortened with ellipses,
  and the named exemplars' forum URLs appear in the bibliography.}
  \label{tab:casestudy}
  \begin{tabular}{p{0.13\columnwidth}p{0.39\columnwidth}p{0.39\columnwidth}}
    \toprule
    Case & AIPR weakness (verbatim) & Human review (verbatim excerpt) \\
    \midrule
    \multicolumn{3}{l}{\emph{Correctly flagged: low score, rejected}}\\[2pt]
    Reject; AIPR 58 & ``Most graph experiments rely on a synthetic three-year timeline built from disconnected Ego4D clips, and the paper does not quantify how much the reported gains depend on that construction.'' & ``Dataset construction lacks transparency \ldots\ Ego4D consists of short clips \ldots\ from 931 different people---how were these transformed into a continuous 3-year personal diary?'' \\[3pt]
    Reject; AIPR 66 & ``The experimental comparison is thin: only ERM and static BIRM are evaluated, with no representative domain-generalization or robust fine-tuning baselines.'' & ``Experiments are insufficient: too few baselines, limited models, and narrow evaluation.'' \ldots\ ``No comparison to CORAL, meta-learning, or other domain generalization methods.'' \\[3pt]
    Reject; AIPR 60 & ``The title and parts of the abstract overclaim relative to the evidence \ldots\ a narrow slice of interpretability tooling.'' & ``Its poor situating within existing literature and \ldots\ major issues in clarity, rigor \ldots\ It makes strong claims about deception and interpretability without referencing key prior studies.'' \\
    \midrule
    \multicolumn{3}{l}{\emph{False negative: high score, rejected}}\\[2pt]
    Reject; AIPR 77 & ``Two claims are framed too strongly for the evidence provided \ldots\ size optimization is said not to hurt performance even though appendix tables show non-trivial drops on harder sets.'' & ``The contributions are not very strong \ldots\ an arguably trivial combination of existing neural circuit-synthesis, feedback-based fine-tuning, and expert-iteration methods.'' \\
    \midrule
    \multicolumn{3}{l}{\emph{False positive: low score, accepted oral}}\\[2pt]
    Oral; AIPR 74 & ``The theory assumes independent updates and ignores optimizer state, yet the paper makes broader optimizer-agnostic and decay-emulation claims; that gap is not bridged.'' & ``Algebra aside, the theory is a bit specious \ldots\ really I think this is `decay-inspired averaging'.'' \emph{(this reviewer rated it 10; accepted oral)} \\[3pt]
    Oral; AIPR 75 & ``The paper's strongest theoretical claim overreaches the proof chain.'' & ``This method also has cryptographic overhead (Eq.\ 13). Hence, the `encryption-free' claim is potentially misleading.'' \emph{(accepted oral)} \\
    \bottomrule
  \end{tabular}
\end{table}

\section{Additional threats to validity}
\label{app:limits}
The main-text limitations (\S\ref{sec:limits}) cover the load-bearing threats.
The additional, lower-order caveats below are recorded here for completeness.

\paragraph{Exclusion (arXiv-twin) bias.}
We exclude submissions whose self-identity step flags them as an already-published
manuscript under a different title, removing an arXiv-twin leakage risk for the
citation audit (\S\ref{sec:data}). Excluding papers on a content-derived signal
could in principle bias the retained sample. We therefore report the headline
discrimination and rating correlation both with and without the exclusion, and
report the excluded set's outcome distribution; the signal of interest is that the
two agree (i.e.\ the exclusion does not manufacture the effect), and the excluded
count is small relative to the cohort.

\paragraph{Submitted vs.\ post-rebuttal.}
We grade the submitted PDF, but reviewer ratings can shift after rebuttal and
revision. The score is computed on the artifact reviewers first saw; any residual
mismatch between that artifact and the final rating adds noise against us rather
than in our favor.

\paragraph{Grounded is not relevant.}
The citation-grounding advantage of the deployed product's search-backed audit
establishes that a recommended reference is a real record returned by the retrieval
tool, not that it is the \emph{most relevant} omission. Groundedness rules out hallucinated
citations; it does not certify editorial judgment about which prior work matters
most, which remains a human call.

\paragraph{Run-to-run variance.}
The score is stochastic across runs but only mildly so: median within-paper SD
$\approx\runSDmedian$ points on the frontier configuration, estimated from
\runSDnPapers{} papers graded three times each (\S\ref{sec:res-value}). Cross-paper
point estimates use each paper's original single run (the pre-registered
single-run design); the variance sub-study's re-runs are an augmentation and do not
enter the headline metrics. The re-grade estimate is computed over the two
consistent-configuration re-runs per paper, excluding run~0, the original frontier
run whose citation audit returned empty (the pinned-100 artifact of
\S\ref{sec:discussion}), so the SD reflects stochastic grading noise rather than
that one-off scoring difference. The same two-re-run estimate, computed identically
for the Direct grader on the same papers, supports the paired comparison: an exact
Wilcoxon signed-rank test on the \relPairedN{} paired within-paper SDs gives
\relPairedP, so the reliability gap is systematic across papers rather than a
median artifact. The variance is small relative to the between-tier separation but
is not zero.

\section{Failure modes in detail}
\label{sec:failure}

This appendix gives the full treatment summarized in \S\ref{sec:res-secondary}. A score
that is useful in deployment must be characterized where it errs as well as where it
succeeds. We examine both directions of error on cohort $H$, and because the venue is
open we read the actual human reviews to explain each.
Table~\ref{tab:casestudy} (Appendix~\ref{app:cases}) places AIPR's stated weakness
beside a verbatim excerpt from the human reviews for a matched set of submissions
spanning the patterns below; we paraphrase here and quote in full there.

\subsection{What the score catches (and why)}
For rejected submissions the score rated low, the in-depth review's stated
weaknesses align closely with the reasons human reviewers gave, and at the level of
the specific defect rather than merely the overall verdict. The recurring,
correctly-flagged patterns, each instantiated in Table~\ref{tab:casestudy}, are:
\begin{itemize}\itemsep1pt
  \item \textbf{Unquantified or synthetic data construction}: AIPR flags that the
  reported gains may hinge on how a synthetic benchmark was built, and the human
  reviewers independently interrogate the same construction.
  \item \textbf{Thin or missing baselines}: AIPR names the absent comparisons
  (e.g.\ ``only ERM and static BIRM''); reviewers name the same gap (``too few
  baselines'').
  \item \textbf{Overclaiming and poor situating}: AIPR flags a gap between the
  title/abstract claims and the evidence, and missing prior work; reviewers lead
  with the identical objection.
\end{itemize}
In each case the model and the reviewers point at the same table, claim, or
omission, which is what makes a low score a usable flag rather than a coincidence.

\subsection{Where the score is wrong}
We report both error classes explicitly, with examples in Table~\ref{tab:casestudy}.
\paragraph{False negatives: rejected work the score rated highly.}
The dangerous case for a triage tool is a weak paper scored high. In the observed
examples these concentrate where the rejection rests on a \emph{contribution-level
judgment} that a competent manuscript surface does not betray:
\ifanonymous for one rejected submission (AIPR~77),\else for the rejected
\citet{reactivesynth2026} (AIPR~77),\fi{} reviewers call the method ``an arguably
trivial combination of existing'' techniques with ``little conceptual novelty,''
while AIPR, scoring a well-written and well-executed paper, places it in the upper
band. A second, structural cause is a defect the system cannot see at all: it reads
the manuscript as written and cannot execute an experiment, reproduce a result, or
re-derive a proof, so where rejection hinges on such a defect a high score is expected
and is not evidence against the bounded claim of \S\ref{sec:results}. Either way, the rate of these errors at the top of the score
range is quantified directly in the reliability curve (Fig.~\ref{fig:validation}b).

\paragraph{False positives: strong work the score underrated.}
The mirror error is accepted work scored low. In the observed examples AIPR's lower
score reflects a \emph{genuine} concern (an over-reaching theoretical claim, or a
theory whose assumptions do not match practice) that a human reviewer often raised
as well, but that the community forgave in light of the contribution: the papers
were accepted as orals on their merits. AIPR underweights a strong contribution when
it carries a real but pardonable gap. These cost a second human look, which is the
intended action anyway, and they bound any use of the score for positive selection.

\subsection{Reading of the failure modes}
The two error classes differ in their consequences for deployment. A false
negative at the \emph{top} of the score range (a weak paper scored high) is the
dangerous case for a triage tool, and we quantify its rate directly in the
reliability curve (Fig.~\ref{fig:validation}b). A false positive (a strong paper
scored low) costs a second human look, which is the intended action anyway. This
asymmetry is what makes the narrow claim, to flag weak work for human attention,
defensible as a human-reviewed triage signal under the studied conditions even
where the score ranks strong work poorly.

\section{Threats to validity and limitations (full)}
\label{sec:limits}

This appendix is the full version of the limitations summarized in \S\ref{sec:discussion}.

\paragraph{Population and ground-truth noise.}
Submissions to a top venue are self-selected above average, so the reject class here
is ``weak \emph{relative to the venue bar},'' not weak in absolute terms; we do not
generalize beyond this population. The accept/reject decision is itself noisy and
partly reflects reviewer disagreement, area-chair effects, and venue fit
\citep{cortes2021inconsistency,beygelzimer2023arbitrary}. We mitigate this by also
validating against the continuous mean reviewer rating (H4), but no analysis here can
exceed the reliability of its ground truth.

\paragraph{Leakage and contamination.}
Beyond verbatim memorization, excluded by construction through the post-cutoff cohort,
the model holds priors on institutions, authors, and topics that could correlate with
outcomes. We address this with the arXiv-before-cutoff sensitivity split
(\S\ref{sec:discussion}), and the question of whether printed identity moves the score
by the separate prestige-perturbation experiment. A fully pre-cutoff contaminated
contrast and a prospective cohort are planned follow-ups, not part of this analysis.
Residual indirect familiarity cannot be fully excluded retrospectively; only the
prospective cohort closes it entirely.

\paragraph{Ground truth may itself be partly machine-generated.}
A growing fraction of submitted peer reviews are now LLM-assisted at the venues we
study \citep{liang2024monitoring}. To the extent the reviewer ratings we validate
against are themselves shaped by language models, our agreement metric partly measures
agreement with other AI systems rather than with unaided human judgment. We cannot
separate the two components with the available data; the effect, if present, inflates
apparent agreement and is a ceiling on the interpretation of H4.

\paragraph{The cheap-model proxy.}
The scale of the headline rests on the bridge (H5), which we measure and gate on
rather than assume. The bridge is strongest globally and weaker exactly where the
deployable claim lives: within the cohort-$M$ bottom quintile the two scores agree at
only $\rho =$~\bridgeLowRhoFull{} and \bridgeOverlap\% of that band recurs in the cohort-$H$
bottom quintile (\S\ref{sec:res-validity}). A low AIPR (GPT-5.4-mini) flag is therefore
best read as a screen to be confirmed on the AIPR (GPT-5.4) score, not an exact
substitute on an individual submission. This is why we report the deployable bottom-flag
(H1) on AIPR (GPT-5.4) directly (\bottomRejectRateFrontier\%): the production claim does
not route through the weak low-end bridge.

\paragraph{The excluded citation dimension.}
In this run the citation audit returned no recommendations and the pipeline scored an
empty audit as a perfect bibliography, so the citation subscore was uninformative (AUROC
\citationAUROC, chance). We exclude it from interpretation; because it is the most lightly
weighted dimension, dropping it leaves discrimination unchanged (AUROC
\aurocDropCitationFull{} vs.\ \aurocFull). Scoring an empty audit as \emph{unknown} and
re-grading under a search-backed configuration is the fix (\S\ref{sec:discussion}).

\paragraph{Single venue and field.}
The primary cohort is one venue in one field, and we do not claim cross-field
generalization here. A planned follow-up grades a full review set at a second,
different-field venue; because venues that publish \emph{rejected} submissions are
rare, that study validates against the continuous reviewer rating, which generalizes
across fields, rather than the reject class, and is reported separately. A
within-family contaminated contrast (a prior ICLR year) is a planned follow-up that
addresses year-to-year stability only.

\paragraph{Defined semantics, fixed weights.}
A score's meaning is fixed by the grading rubric, with each dimension defined by
explicit criteria, so it is an anchored construct, and the overall is a fixed weighted
mean (Eq.~\ref{eq:overall}) set on substantive grounds and deliberately \emph{not} fit
to the outcome. Holding the weights label-independent is what licenses the validity
test. We validate that submissions \emph{ordered} by the rubric track the human signal
(discrimination and rank agreement); we do \emph{not} calibrate the absolute level to
an acceptance probability, a distinct venue-specific exercise we leave to future work.

\paragraph{Reproducibility with a proprietary scorer.}
The scored dataset, the decision and rating labels, the analysis code, and this paper
are released, so every figure, statistic, and the entire score-to-outcome analysis
reproduces independently. The grading rubric, prompts, and model configuration are
proprietary: the manuscript-to-score function is audited through its outputs, and can
be exercised on new manuscripts via the system, rather than re-implemented. This is a
defensible posture for validating a deployed instrument, though weaker than full
independent reproducibility of the scorer itself. Because the design was frozen and
pre-registered before unblinding, this openness is what lets a single author's
validation be checked rather than taken on trust. Additional minor threats
(post-rebuttal rating drift, citation relevance vs.\ groundedness, run-to-run variance)
are detailed in Appendix~\ref{app:limits}.

\paragraph{Sample-vs-eligible representativeness and per-review sensitivities.}
The population-boundary ledger (Appendix~\ref{app:robust}, Table~\ref{tab:popboundary})
reports counts and exclusion reasons for every eligible submission, but a
covariate-level comparison of the graded cohort to the full eligible population is not
possible because manuscript covariates were not extracted for the ungraded submissions.
The released two-CSV data carries only the mean reviewer rating, so per-review
sensitivities (a human-review reliability ceiling and alternate rating aggregations)
are deferred to future work and available on request; we do not synthesize per-review
ratings to fill the gap.

\paragraph{What we do not claim.}
We do not claim the score predicts acceptance probability, ranks strong papers, or
should be used for positive selection or any consequential decision without a human in
the loop. The validated use is narrow: the action we validate is prioritization for
human attention, not rejection. The deployment-boundary, independence, and
proprietary-scorer audit commitments this posture rests on are collected in
Appendix~\ref{app:posture}.


\end{document}